\pdfoutput=1

\documentclass[11pt]{article}

\usepackage[table,xcdraw]{xcolor}         

\usepackage[]{ACL2023}

\usepackage{times}
\usepackage{latexsym}

\usepackage[T1]{fontenc}

\usepackage[utf8]{inputenc}

\usepackage{microtype}

\usepackage{hyperref}       
\usepackage{url}            
\usepackage{booktabs}       
\usepackage{amsfonts}       
\usepackage{nicefrac}       
\usepackage{microtype}      
\usepackage{amsmath, amssymb}
\usepackage{multirow}
\usepackage{graphicx, float, subfig}
\usepackage{adjustbox}
\usepackage{wrapfig}
\usepackage{listings}
\usepackage{amsthm}
\usepackage{enumitem}
\theoremstyle{definition}
\newtheorem{definition}{Definition}[section]
\newcommand*{\affaddr}[1]{#1} 

\makeatletter
\DeclareRobustCommand\bigop[1]{%
  \mathop{\vphantom{\sum}\mathpalette\bigop@{#1}}\slimits@
}
\newcommand{\bigop@}[2]{%
  \vcenter{%
    \sbox\z@{$#1\sum$}%
    \hbox{\resizebox{\ifx#1\displaystyle.9\fi\dimexpr\ht\z@+\dp\z@}{!}{$\m@th#2$}}%
  }%
}
\makeatother

\newcommand{\red}[1]{{\color[HTML]{C8686D}#1}}
\newcommand{\green}[1]{{\color[HTML]{55A15D}#1}}

\newcommand{\impro}[1]{{\hspace{0.05cm}{\color[HTML]{32CB00}\textbf{(+#1)}}}}

\DeclareMathOperator*{\eop}{\mathbb{E}}

\definecolor{tabhighlight}{HTML}{e5e5e5}
\definecolor{grey}{RGB}{128,138,135}

\usepackage{inconsolata}

\title{Delving into the Openness of CLIP}

\author{%
  Shuhuai Ren\quad Lei Li\quad Xuancheng Ren\quad Guangxiang Zhao\quad Xu Sun \\
  \affaddr{National Key Laboratory for Multimedia Information Processing, \\ School of Computer Science, Peking University}\\
  \texttt{\{shuhuai\_ren,lilei\}@stu.pku.edu.cn} \\
  \texttt{\{renxc,zhaoguangxiang,xusun\}@pku.edu.cn} \\
}

\begin{document}
\maketitle
\begin{abstract}
Contrastive Language-Image Pre-training (CLIP) formulates image classification as an image-to-text matching task, i.e., matching images to the corresponding natural language descriptions instead of discrete category IDs. This allows for open-vocabulary visual recognition, where the model can recognize images from an open class set (also known as an open vocabulary) in a zero-shot manner. However, evaluating the openness of CLIP-like models is challenging, as the models are open to arbitrary vocabulary in theory, but their accuracy varies in practice. 
To address this, we resort to an incremental perspective to assess the openness through vocabulary expansions, and define \emph{extensibility} to measure a model's ability to handle novel classes. Our evaluation shows that CLIP-like models are not truly open, and their performance deteriorates as the vocabulary expands. 
We further dissect the feature space of CLIP from the perspectives of representation alignment and uniformity. 
Our investigation reveals that the overestimation of openness is due to confusion among competing text features, rather than a failure to capture the similarity between image features and text features of novel classes. We hope that our investigation and analysis will facilitate future research on the CLIP openness issue.\footnote{Our code is available at \url{https://github.com/lancopku/clip-openness}}
\end{abstract}

\section{Introduction}
An intrinsically open mechanism for visual recognition~\cite{Deng2009ImageNetAL, He2016DeepRL} has always been a shared goal in the computer vision community~\cite{Scheirer2013TowardOS, Geng2021RecentAI, Bendale2015TowardsOW}. 
This mechanism requires models to maintain flexibility to cope with the scaling of the recognition target, where both input images and the corresponding classes will dynamically expand according to actual needs. 
For example, in medical diagnosis~\cite{Razzak2017DeepLF}, new diseases emerge constantly, and in e-commerce, new categories of products appear daily~\cite{Xu2019OpenworldLA}, which cannot be predefined in a finite, fixed class set. 

Faced with the challenging task of open-world recognition, Contrastive Language-Image Pre-training (CLIP)~\cite{Radford2021LearningTV} and its open-vocabulary learning paradigm demonstrate superiority over traditional supervised classifiers~\cite{He2016DeepRL, Dosovitskiy2021AnII}. 
CLIP pre-trains a vision-language model on web-scale collections of image-text pairs, learning semantic alignment between images and corresponding textual descriptions. 
During inference, it formulates image classification as an image-to-text matching task, where the set of class names serves as a vocabulary, and textual prompts like "\texttt{a photo of a [CLASSNAME]}" are curated as class descriptions for images. By varying the \texttt{[CLASSNAME]} placeholder and computing the similarity between class descriptions and images, CLIP can identity the most suitable class name and predict it as the target class. 
This approach allows CLIP to operate with arbitrary vocabularies and adapt to novel classes by expanding the vocabulary, enabling zero-shot inference for new input images and classes. 

Nevertheless, previous evaluation protocols for CLIP models only assess their accuracy on static, closed vocabularies from downstream datasets, leaving their actual performance on open tasks in the shadows~\cite{Radford2021LearningTV}. 
In this work, we delve into openness, the intriguing yet under-explored property in CLIP-like models~\cite{Li2021SupervisionEE, Mu2021SLIPSM, Yao2021FILIPFI, Zhou2021LearningTP}, and present a novel protocol for evaluating the openness from an incremental view. 
Specifically, we define a metric of \textbf{extensibility} to measure a model's ability to handle new visual concepts through vocabulary expansion. 
Different from previous metrics, our metric explicitly models the dynamics of the real open world, and formulates the empirical risk of CLIP when new vocabularies incrementally emerge. 
Additionally, we define a metric of \textbf{stability} to explore how stable the model's predictions are for old classes when new classes are introduced, which provides a tool to analyze the compatibility between different classes. 

Using our protocol, we conduct a systematic and comprehensive evaluation of CLIP-like models. 
Our experimental results based on extensibility show that CLIP and its variants have a significant drop in accuracy as the vocabulary size increases. For example, CLIP (RN101) on CIFAR100 experiences a $12.9\%$ drop in accuracy when the vocabulary size expands from $5$ to $100$. This indicates that the limited zero-shot capability of CLIP-like models is inadequate for supporting their deployment in the open world. 
What's worse, through an analysis of the prediction shift during vocabulary expansion, we find that the performance of CLIP can be dramatically reduced by adding only three adversarial class names into the vocabulary, exposing the model's poor stability and security risks. 
Furthermore, we investigate the representation space of CLIP-like models via three metrics: margin, inter-modal alignment, and intra-modal uniformity.
Our results show that the small margin between positive and negative class descriptions leads to prediction shifting when competing class features appear. Therefore, enforcing the distinguishability of class features increases the margin and improves the stability of these models. 

In summary, our contribution is threefold: \textbf{First}, to the best of our knowledge, we are the first to systematically quantify the openness of CLIP, for which we design the evaluation protocol and two indicators of extensibility and stability. 
\textbf{Second}, we conduct extensive experiments on CLIP-like models based on our protocol and find that their openness is overestimated and their performance declines as the vocabulary expands. \textbf{Finally}, we analyze the feature space of CLIP from the perspectives of representation alignment and uniformity, observing that the uniformity of the textual space is critical for better extensibility. 

\section{Related work}
\paragraph{Contrastive language-image pre-training and open-vocabulary learning.}
CLIP~\cite{Radford2021LearningTV} introduces the paradigm of open-vocabulary learning and learns transferable visual models from natural language supervision. The CLIP model consists of an image encoder and a text encoder, which are utilized to encode image-text pairs into a joint feature space for learning the semantic alignment of vision and language. The paired images and texts are pulled together in the feature space, while the others with dissimilar semantics are pushed apart via a contrastive loss. 
After pre-training on large-scale image-text pairs, CLIP is able to map images to their corresponding language descriptions, which makes visual recognition generalize in the wild. 
Recent studies further improve CLIP by using more pre-training data~\cite{Jia2021ScalingUV}, incorporating self-supervision~\cite{Mu2021SLIPSM}, fine-grained supervision~\cite{Yao2021FILIPFI}, and widespread supervision~\cite{Li2021SupervisionEE} to pre-training. 
Another line of recent studies~\cite{li2021align, wang2022unifying, yu2022coca, alayrac2022flamingo} adopts seq2seq generation instead of contrastive discrimination framework to achieve open-vocabulary recognition. We leave the investigation of their extensibility for future work. 

\paragraph{Open Set and Open-World Visual Recognition.}
Open Set Recognition (OSR)~\cite{Scheirer2013TowardOS, Geng2021RecentAI} and Open World Recognition (OWR)~\cite{Bendale2015TowardsOW} are paradigms aiming to cope with input images from novel classes during inference. 
OSR requires classifiers to identify images that have not been introduced during training as ``unknown''. 
While OWR raises higher demands, models are supposed to incrementally extend and retrain the multi-class classifier as the unknowns are labeled as additional training data. 
Contrary to the above research, CLIP-based Open-vocabulary Recognition (OVR) aims to identify novel classes in a zero-shot manner by using natural language representations of categories instead of discrete label IDs. This allows CLIP to directly synthesize textual descriptions of novel classes for matching, eliminating the need for relabeling additional training data and re-training the entire model. A more detailed comparison of OSR, OWR, and OVR can be found in Appendix~\ref{sec:related-work-comparison}.

\begin{figure*}[tbp]
    \centering
    \includegraphics[width=.95\textwidth]{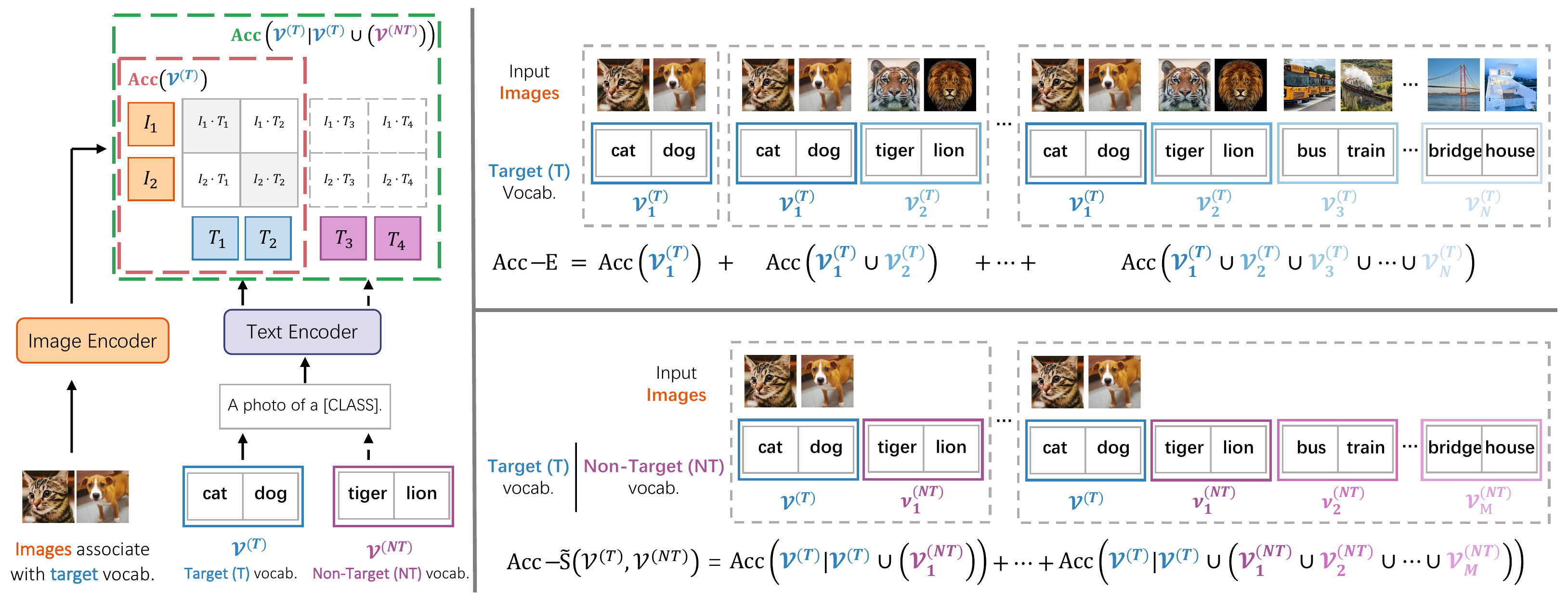}
\caption{\textbf{Left}: the \red{original accuracy} of CLIP with target vocabulary (Eq.\eqref{eq:CLIP-acc}) and the \green{conditional accuracy} of CLIP with non-target vocabulary (Eq.\eqref{eq:CLIP-condition-acc}). In the latter, the classes from the non-target vocabulary are involved as distractors for input images restricted in the target vocabulary. \textbf{Upper right}: calculation of Acc-E (Eq.\eqref{eq:acc-e}). It measures the extensibility of models when recognition targets, including both classes and the associated input images, are scaling simultaneously. \textbf{Bottom right}: calculation of Acc-S (Eq.\eqref{eq:acc-s}), a sub-problem introduced by Acc-E. It measures the prediction stability on the images from the target vocabulary as the distractors from the non-target vocabularies are incorporated incrementally.}
\label{fig:overview}
\end{figure*}

\section{Openness, Extensibility, and Stability}
\label{sec:openness-extensibility-stability}
In this section, we first review CLIP's visual recognition paradigm and demonstrate how it realizes open-vocabulary image classification through vocabulary expansion~(\textsection~\ref{subsec:openness}). To quantify the actual performance of CLIP-like models as the vocabulary expands, we define the metric of extensibility and propose a systematical evaluation protocol~(\textsection~\ref{subsec:extensibility}). The experimental results and further analysis reveal that, as the vocabulary expands, CLIP's predictions become unstable and prone to drift towards newly introduced competing class descriptions, which limits its extensibility and poses a huge security risk when deployed in real-world applications~(\textsection~\ref{subsec:stability}).

\subsection{Openness of CLIP}
\label{subsec:openness}
CLIP~\cite{Radford2021LearningTV} models image classification as an image-to-text matching task. Formally, let $f$ be the CLIP model, $f_T$ and $f_I$ be the text and image encoders in CLIP, respectively. The CLIP model takes an image $x$ and a \textit{target vocabulary} $\mathcal{V}^{(T)}=\left\{ w_i \right\}$ of the class names $w_i$ as inputs, and predicts the image label as:
\begin{equation*}
\small
\begin{split}
f\left( x, \mathcal{V}^{(T)} \right) &= \arg \max_{i} P\left( y=i \mid x \right) \\
&= \arg \max_{i} \frac{e ^{  \operatorname{sim}( f_T( t_i), f_I(x) )  }}{\sum\limits_{j=1}^{|\mathcal{V}^{(T)}|} e^{  \operatorname{sim}( f_T(t_j ), f_I(x) ) }},
\end{split}
\label{eq:CLIP-y}
\end{equation*}
where $t_i$ is the textual description of the class name $w_i$ in a prompt format, e.g., ``\texttt{a photo of a $w_i$}'', and $\operatorname{sim}(\cdot, \cdot)$ denotes cosine similarity. 
Such a modeling paradigm can realize open-world image classification in theory by extending the target vocabulary $\mathcal{V}^{(T)}$ to arbitrary degrees. 
However, in most previous work~\cite{Radford2021LearningTV, Li2021SupervisionEE, Mu2021SLIPSM, Yao2021FILIPFI, Zhou2021LearningTP}, CLIP is evaluated with a fixed vocabulary $\mathcal{V}^{(T)}$ of a downstream dataset $\mathcal{D}^{(T)}$: 
\begin{equation}
\small
\text{Acc}\left( \mathcal{V}^{(T)} \right) = \frac{1}{|\mathcal{D}^{(T)}|} \sum_{(x,y)\in \mathcal{D}^{(T)}} \mathbb{I} \Big( f\left( x, \mathcal{V}^{(T)} \right) = y \Big),
\label{eq:CLIP-acc}
\end{equation}
where $|\mathcal{D}^{(T)}|$ is the size of the dataset and $\mathbb{I}(\cdot)$ is the indicator function. 
This vanilla evaluation setting, utilizing restricted input images and classes, falls short for open recognition tasks. It fails to consider the dynamic expansion of vocabulary during inference and, as a result, cannot accurately reflect CLIP's openness in real-world scenarios where the number of classes may increase.

\begin{table*}[tbp]
\centering
\setlength{\tabcolsep}{3pt}
\begin{adjustbox}{max width=\textwidth}
\begin{tabular}{@{}lccccccccccccccc@{}}
\toprule
                        & \multicolumn{5}{c}{CIFAR100}                                                                                                   & \multicolumn{5}{c}{ImageNet (Entity13)}                                                                                        & \multicolumn{5}{c}{ImageNet (Living17)}                                                                                        \\ \cmidrule(l){2-6} \cmidrule(l){7-11} \cmidrule(l){12-16} 
                        &                         & \multicolumn{2}{c}{\cellcolor[HTML]{E2F0D9}Extensibility}                     & \multicolumn{2}{c}{\cellcolor[HTML]{DAE3F3}Stability}                         &                         & \multicolumn{2}{c}{\cellcolor[HTML]{E2F0D9}Extensibility}                     & \multicolumn{2}{c}{\cellcolor[HTML]{DAE3F3}Stability}                         &                         & \multicolumn{2}{c}{\cellcolor[HTML]{E2F0D9}Extensibility}                     & \multicolumn{2}{c}{\cellcolor[HTML]{DAE3F3}Stability}                         \\ \cmidrule(lr){3-4} \cmidrule(lr){5-6} \cmidrule(lr){8-9} \cmidrule(lr){10-11} \cmidrule(lr){13-14} \cmidrule(lr){15-16} 
\multirow{-3}{*}{Model} & \multirow{-2}{*}{Acc-C} & \cellcolor[HTML]{E2F0D9}Acc-E         & \cellcolor[HTML]{E2F0D9}$\Delta$      & \cellcolor[HTML]{DAE3F3}Acc-S         & \cellcolor[HTML]{DAE3F3}$\Delta$      & \multirow{-2}{*}{Acc-C} & \cellcolor[HTML]{E2F0D9}Acc-E         & \cellcolor[HTML]{E2F0D9}$\Delta$      & \cellcolor[HTML]{DAE3F3}Acc-S         & \cellcolor[HTML]{DAE3F3}$\Delta$      & \multirow{-2}{*}{Acc-C} & \cellcolor[HTML]{E2F0D9}Acc-E         & \cellcolor[HTML]{E2F0D9}$\Delta$      & \cellcolor[HTML]{DAE3F3}Acc-S         & \cellcolor[HTML]{DAE3F3}$\Delta$      \\ \midrule
CLIP (RN101)            & 68.3                    & \cellcolor[HTML]{E2F0D9}55.4          & \cellcolor[HTML]{E2F0D9}-12.9         & \cellcolor[HTML]{DAE3F3}54.9          & \cellcolor[HTML]{DAE3F3}-13.4         & 80.4                    & \cellcolor[HTML]{E2F0D9}77.4          & \cellcolor[HTML]{E2F0D9}-3.0          & \cellcolor[HTML]{DAE3F3}77.3          & \cellcolor[HTML]{DAE3F3}-3.1          & 77.6                    & \cellcolor[HTML]{E2F0D9}74.5          & \cellcolor[HTML]{E2F0D9}-3.1          & \cellcolor[HTML]{DAE3F3}74.4          & \cellcolor[HTML]{DAE3F3}-3.2          \\
CLIP (ViT-B/32)         & 78.0                    & \cellcolor[HTML]{E2F0D9}69.6          & \cellcolor[HTML]{E2F0D9}-8.4          & \cellcolor[HTML]{DAE3F3}68.9          & \cellcolor[HTML]{DAE3F3}-9.1          & 80.8                    & \cellcolor[HTML]{E2F0D9}78.0          & \cellcolor[HTML]{E2F0D9}-2.8          & \cellcolor[HTML]{DAE3F3}77.8          & \cellcolor[HTML]{DAE3F3}-3.0          & 78.0                    & \cellcolor[HTML]{E2F0D9}74.4          & \cellcolor[HTML]{E2F0D9}-3.6          & \cellcolor[HTML]{DAE3F3}75.0          & \cellcolor[HTML]{DAE3F3}-3.0          \\
CLIP (ViT-B/16)         & \textbf{79.7}           & \cellcolor[HTML]{E2F0D9}\textbf{72.6} & \cellcolor[HTML]{E2F0D9}\textbf{-7.1} & \cellcolor[HTML]{DAE3F3}\textbf{72.0} & \cellcolor[HTML]{DAE3F3}\textbf{-7.7} & \textbf{83.5}           & \cellcolor[HTML]{E2F0D9}\textbf{81.1} & \cellcolor[HTML]{E2F0D9}\textbf{-2.4} & \cellcolor[HTML]{DAE3F3}\textbf{81.0} & \cellcolor[HTML]{DAE3F3}\textbf{-2.5} & \textbf{79.5}           & \cellcolor[HTML]{E2F0D9}\textbf{77.9} & \cellcolor[HTML]{E2F0D9}\textbf{-1.6} & \cellcolor[HTML]{DAE3F3}\textbf{77.6} & \cellcolor[HTML]{DAE3F3}\textbf{-1.9} \\ \midrule
SLIP (ViT-B/16)         & 63.9                    & \cellcolor[HTML]{E2F0D9}51.1          & \cellcolor[HTML]{E2F0D9}-12.8         & \cellcolor[HTML]{DAE3F3}50.4          & \cellcolor[HTML]{DAE3F3}-13.5         & 65.7                    & \cellcolor[HTML]{E2F0D9}62.3          & \cellcolor[HTML]{E2F0D9}-3.4          & \cellcolor[HTML]{DAE3F3}62.0          & \cellcolor[HTML]{DAE3F3}-3.7          & 65.7                    & \cellcolor[HTML]{E2F0D9}62.6          & \cellcolor[HTML]{E2F0D9}-3.1          & \cellcolor[HTML]{DAE3F3}62.5          & \cellcolor[HTML]{DAE3F3}-3.2          \\
DeCLIP (ViT-B/32)       & \textbf{78.7}           & \cellcolor[HTML]{E2F0D9}\textbf{70.8} & \cellcolor[HTML]{E2F0D9}\textbf{-7.9} & \cellcolor[HTML]{DAE3F3}\textbf{70.4} & \cellcolor[HTML]{DAE3F3}\textbf{-8.3} & \textbf{81.9}           & \cellcolor[HTML]{E2F0D9}\textbf{79.2} & \cellcolor[HTML]{E2F0D9}\textbf{-2.7} & \cellcolor[HTML]{DAE3F3}\textbf{79.1} & \cellcolor[HTML]{DAE3F3}\textbf{-2.8} & \textbf{82.1}           & \cellcolor[HTML]{E2F0D9}\textbf{80.2} & \cellcolor[HTML]{E2F0D9}\textbf{-1.9} & \cellcolor[HTML]{DAE3F3}\textbf{80.0} & \cellcolor[HTML]{DAE3F3}\textbf{-2.1} \\ \midrule
PE (ViT-B/32)           & 78.3                    & \cellcolor[HTML]{E2F0D9}70.3          & \cellcolor[HTML]{E2F0D9}-8.0          & \cellcolor[HTML]{DAE3F3}69.9          & \cellcolor[HTML]{DAE3F3}-8.4          & 81.9                    & \cellcolor[HTML]{E2F0D9}79.4          & \cellcolor[HTML]{E2F0D9}-2.5          & \cellcolor[HTML]{DAE3F3}79.2          & \cellcolor[HTML]{DAE3F3}-2.7          & 78.7                    & \cellcolor[HTML]{E2F0D9}76.0          & \cellcolor[HTML]{E2F0D9}-2.7          & \cellcolor[HTML]{DAE3F3}75.8          & \cellcolor[HTML]{DAE3F3}-2.9          \\
PE (ViT-B/16)           & \textbf{79.6}           & \cellcolor[HTML]{E2F0D9}\textbf{72.6} & \cellcolor[HTML]{E2F0D9}\textbf{-7.0} & \cellcolor[HTML]{DAE3F3}\textbf{72.0} & \cellcolor[HTML]{DAE3F3}\textbf{-7.6} & \textbf{85.3}           & \cellcolor[HTML]{E2F0D9}\textbf{83.2} & \cellcolor[HTML]{E2F0D9}\textbf{-2.1} & \cellcolor[HTML]{DAE3F3}\textbf{83.1} & \cellcolor[HTML]{DAE3F3}\textbf{-2.2} & \textbf{79.6}           & \cellcolor[HTML]{E2F0D9}\textbf{78.2} & \cellcolor[HTML]{E2F0D9}\textbf{-1.4} & \cellcolor[HTML]{DAE3F3}\textbf{78.0} & \cellcolor[HTML]{DAE3F3}\textbf{-1.6} \\ \midrule
\textcolor{grey}{CoOp (ViT-B/16)}         & \textcolor{grey}{83.6}                    & \cellcolor[HTML]{E2F0D9}\textcolor{grey}{76.9}          & \cellcolor[HTML]{E2F0D9}\textcolor{grey}{-6.7}          & \cellcolor[HTML]{DAE3F3}\textcolor{grey}{76.7}          & \cellcolor[HTML]{DAE3F3}\textcolor{grey}{-6.9}          & \textcolor{grey}{87.5}                    & \cellcolor[HTML]{E2F0D9}\textcolor{grey}{85.3}          & \cellcolor[HTML]{E2F0D9}\textcolor{grey}{-2.2}          & \cellcolor[HTML]{DAE3F3}\textcolor{grey}{85.5}          & \cellcolor[HTML]{DAE3F3}\textcolor{grey}{-2.0}          & \textcolor{grey}{82.7}                    & \cellcolor[HTML]{E2F0D9}\textcolor{grey}{82.6}          & \cellcolor[HTML]{E2F0D9}\textcolor{grey}{-0.1}          & \cellcolor[HTML]{DAE3F3}\textcolor{grey}{81.3}          & \cellcolor[HTML]{DAE3F3}\textcolor{grey}{-1.4}          \\ \bottomrule
\end{tabular}
\end{adjustbox}
\caption{Extensibility and stability of CLIP-like models on CIFAR100 and ImageNet datasets. 
$\Delta$ refers to the decline of Acc-E/Acc-S compared to Acc-C (\%). All models exhibit a clear drop in performance as the openness of tasks increases. PE denotes Prompt Ensemble. CoOp requires fine-tuning with the additional training data in downstream datasets (16-shot for all classes), which can be viewed as the upper bound of other zero-shot models.}
\label{tb:stability-extensibility}
\end{table*}

\subsection{Quantifying extensibility for open world}
\label{subsec:extensibility}
To quantify the model's capability in dealing with newly emerged recognition targets, we propose an evaluation protocol and define a metric of extensibility based on vocabulary expansion. Concretely, we incrementally expand the vocabulary $\mathcal{V}^{(T)}$ in Eq.\eqref{eq:CLIP-acc} by introducing new classes and their associated input images, then evaluate the accuracy after each expansion. 
These accuracy values reflect the model's dynamic performance as openness increases, and the expected average of these values is defined as the model's extensibility. 
In practice, we achieve this expansion by incrementally unioning $N$ disjoint target vocabularies\footnote{Since $\mathcal{V}^{(T)}$ is bound with $\mathcal{D}^{(T)}$ in Eq.\eqref{eq:CLIP-acc}, expanding the target vocabulary also implies expanding $\mathcal{D}^{(T)}$ (including input images and their labels) at the same time, which we omit for brevity.} as shown in the upper right panel of Figure~\ref{fig:overview}.

\begin{definition}[Extensibility]
\label{definition:extensibility}
Given $N$ disjoint target vocabularies $\{ \mathcal{V}^{(T)}_{1}, \cdots, \mathcal{V}^{(T)}_N \}$, we denote the set of all possible permutations of these vocabularies as $\mathcal{S}_{N}$, and $\mathcal{V}^{(T)}_{s_i}$ as the $i^{(th)}$ vocabulary in a permutation $s \in \mathcal{S}_{N}$.
When we union the $i^{(th)}$ vocabulary with the previous $i-1$ vocabularies, we achieve a vocabulary expansion and obtain $\mathcal{V}^{(T)}_{s_1} \cup \cdots \cup \mathcal{V}^{(T)}_{s_i}$. The extensibility refers to the averaged classification accuracy across $N$ incremental expansions as $i$ increases from $1$ to $N$:
\begin{equation}
\small
\text{Acc-E} = \underset{s\in \mathcal{S}_{N}}{\mathbb{E}} \frac{1}{N}\sum_{i=1}^{N} \text{Acc}\left( \mathcal{V}^{(T)}_{s_1} \cup \cdots \cup \mathcal{V}^{(T)}_{s_i} \right).
\label{eq:acc-e}
\end{equation}
\end{definition}

\paragraph{Experimental settings}
We evaluate the extensibility of CLIP and its variants, including DeCLIP~\cite{Li2021SupervisionEE}, SLIP~\cite{Mu2021SLIPSM}, Prompt Ensemble~\cite{Radford2021LearningTV}, CoOp~\cite{Zhou2021LearningTP}, on the CIFAR100~\cite{Krizhevsky2009LearningML} and ImageNet~\cite{Deng2009ImageNetAL} datasets. 
Non-matching methods~\cite{Gao2021CLIPAdapterBV,  Zhang2021TipAdapterTC, Wortsman2021RobustFO}, such as linear probing, are NOT included since they train a classifier with \textit{finite} class vectors, and thus are not suitable for class scaling in operation. 
To construct the vocabulary, we leverage the underlying superclass-class hierarchical structure of the two datasets~\cite{Krizhevsky2009LearningML, Santurkar2021BREEDSBF} by grouping classes that belong to the same superclass into a vocabulary. 
Accordingly, CIFAR100 has $20$ vocabularies, each with $5$ classes. For ImageNet, we utilize two superclass-class structures~\cite{Santurkar2021BREEDSBF}: Entity13 and Living17. The former has $13$ vocabularies, each with $20$ classes, while the latter has $17$ vocabularies, each with 4 classes. 
Tables in the Appendix~\ref{sec:superclass-class-hierarchy} list all the vocabularies in the two datasets. 
For each dataset, we calculate Acc-C, the averaged classification accuracy across all single vocabularies, based on Eq.\eqref{eq:CLIP-acc}:
\begin{equation}
\small
\text{Acc-C} = \frac{1}{N}\sum_{i=1}^{N} \text{Acc}\left( \mathcal{V}^{(T)}_i \right).
\label{eq:acc-c}
\end{equation}
It represents the original model performance on \textit{closed} vocabularies. 
To calculate the expectation in Acc-E, we sample $100\times N$ permutations for $N$ vocabularies and take the average. 

\begin{figure*}[tbp]
  \centering
  \subfloat[Acc-S drops as non-target vocabulary extends (\emph{Insects} as target vocabulary).]{
    \label{sfig:acc-s-insects}
    \includegraphics[width=.4\textwidth]{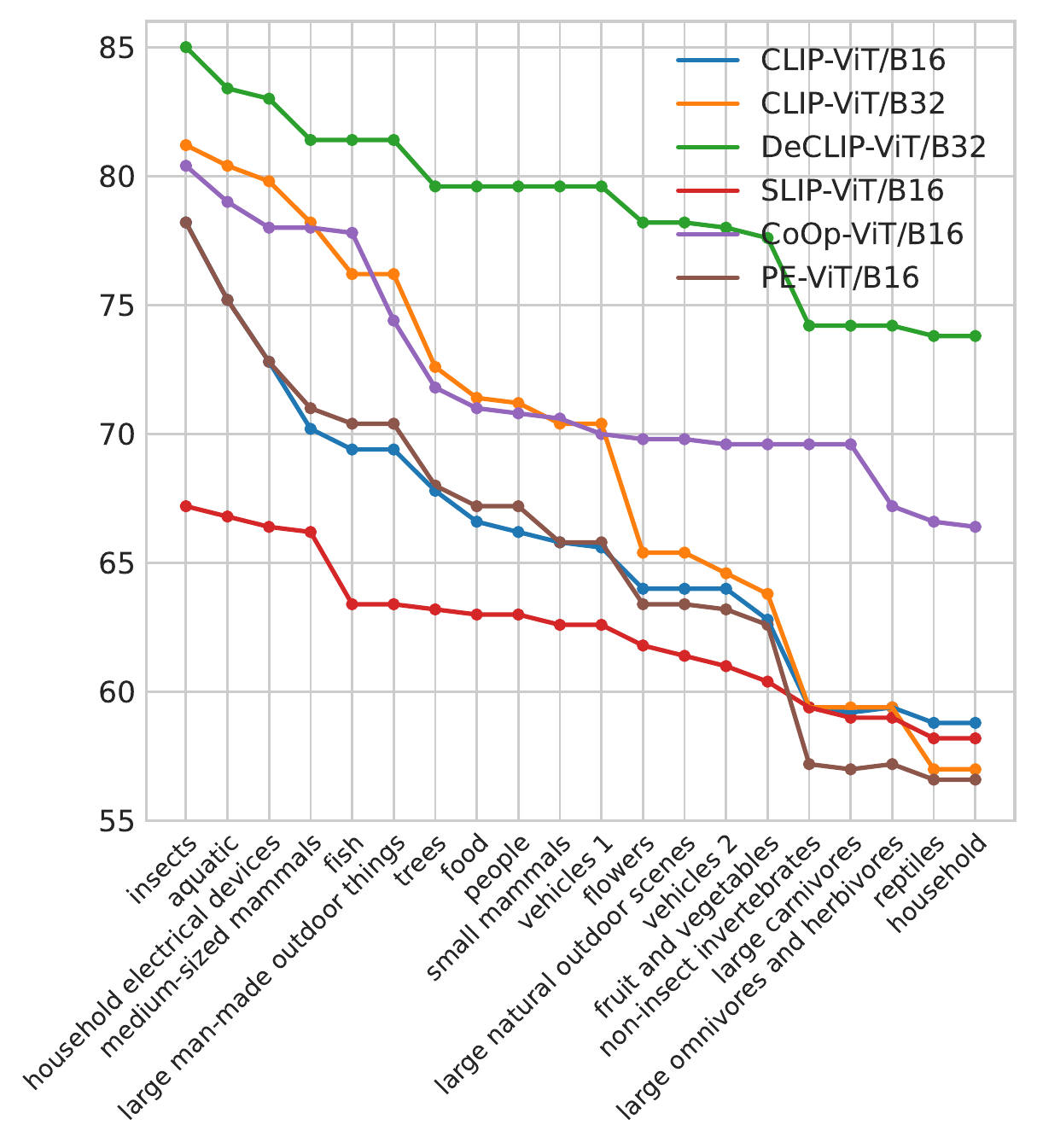}} \hspace{1em}
  \subfloat[Difference between Acc-C and Acc-S of CLIP (ViT-B/32) on different groups.]{
    \label{sfig:acc-o-acc-s}
    \includegraphics[width=.5\textwidth]{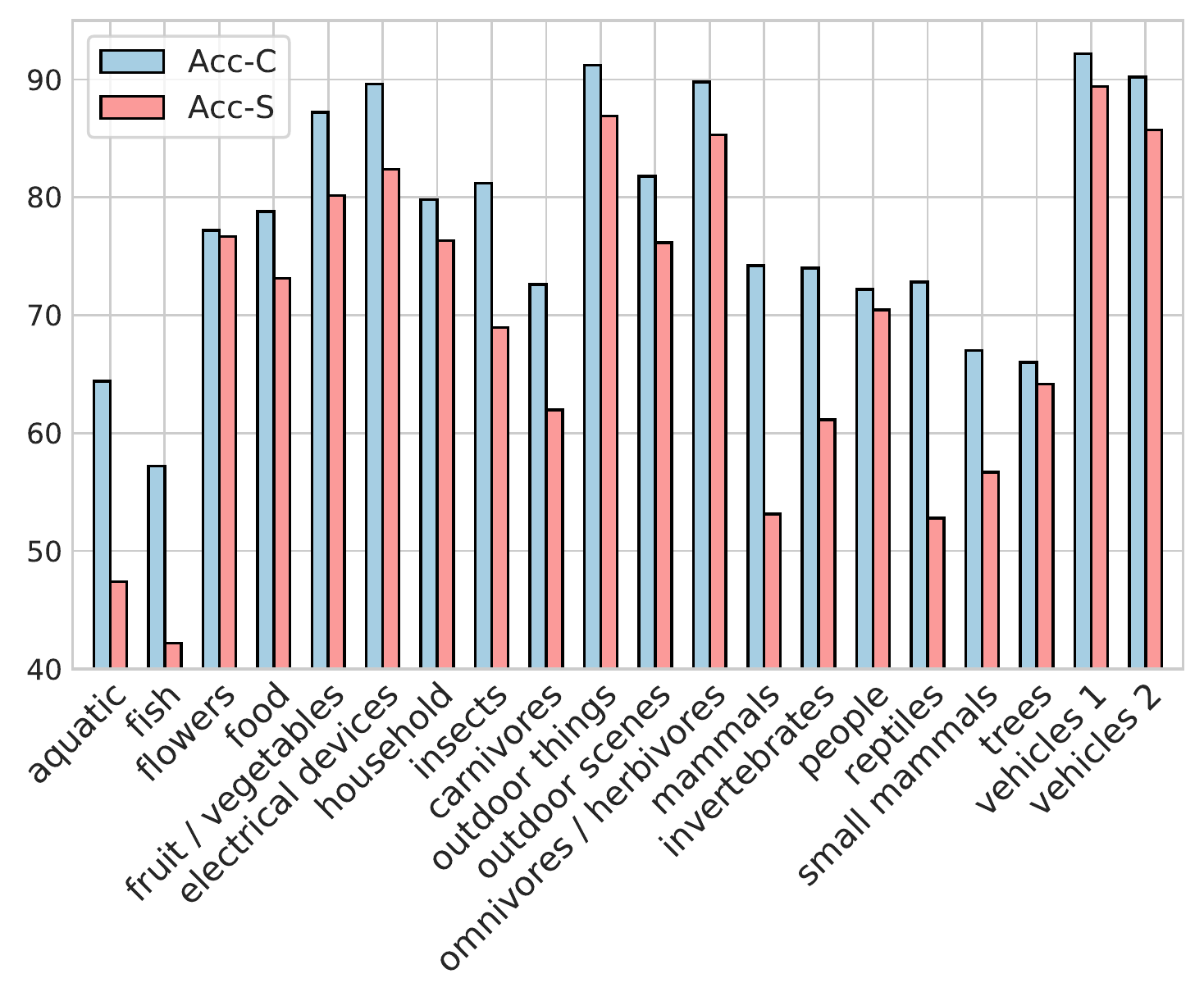}}
  \caption{Acc-C and Acc-S (\%) of CLIP and its variants on CIFAR100. The horizontal axis represents the extended non-target vocabularies in order. PE refers to Prompt Ensemble.}
  \label{fig:acc-s}
\end{figure*}

\paragraph{Results}

As shown in Table~\ref{tb:stability-extensibility},
all models exhibit a clear drop in performance as the vocabulary expands. The accuracy of CLIP (RN101) after vocabulary expansion (Acc-E) sharply decreases by $12.9\%$ compared to the accuracy on closed vocabulary (Acc-C). The performance on the data splits in ImageNet is relatively better, with an average decline of $2.7\%$. 
Appendix~\ref{sec:dataset-level-expansion} provides results of expansion at the dataset level, where the expanded vocabularies are from five other datasets. These results show a more dramatic decline of an average of $15.3\%$ on generic dataset expansion.
It demonstrate that \textbf{the openness of CLIP-like models is overestimated under the vanilla evaluation mechanism}.
Besides, there are some interesting findings: 
\textbf{(1)} From the perspective of pre-training, introducing a stronger vision backbone (ViT~\cite{Dosovitskiy2021AnII} vs. ResNet~\cite{He2016DeepRL}), widespread supervision (DeCLIP~\cite{Li2021SupervisionEE} vs. CLIP), and more pre-training data (CLIP vs. SLIP~\cite{Mu2021SLIPSM}) can improve the extensibility of models on open tasks. 
\textbf{(2)} During inference, the performance of CLIP can be boosted by ensembling different prompts. 
\textbf{(3)} The most extensible results are obtained by CoOp~\cite{Zhou2021LearningTP}, which performs prompt tuning on all classes of CIFAR100 and ImageNet. However, the prompt tuning method utilizes the additional category information and training data, which cannot be applied to real-world open tasks.

\subsection{Stability during vocabulary expansion}
\label{subsec:stability}
As the vocabulary expansion introduces new classes incrementally, some images belonging to previous vocabularies may be incorrectly predicted as new classes, resulting in a drop in accuracy and poor extensibility. 
To analyze the prediction stability of CLIP during vocabulary expansion, we introduce the \textit{non-target classes}. They do NOT correspond to any input images, and only serving as distractors for the target classes. Based on it, we define conditional classification accuracy as: 
\begin{equation}
\small
\begin{split}
& \text{Acc}\left( \mathcal{V}^{(T)} \middle| \mathcal{V}^{(T)} \cup \mathcal{V}^{(NT)} \right) \\
& = \frac{1}{|\mathcal{D}^{(T)}|} \sum_{(x,y)\in \mathcal{D}^{(T)} } \mathbb{I} \left( f\left( x, \mathcal{V}^{(T)}\cup \mathcal{V}^{(NT)} \right) = y \right),
\label{eq:CLIP-condition-acc}
\end{split}
\end{equation}
where $\mathcal{V}^{(NT)}$ is the \textit{non-target vocabulary}, i.e., the vocabulary of non-target classes. 
The conditional accuracy is depicted in the left panel of Figure~\ref{fig:overview}.
In Eq.\eqref{eq:CLIP-condition-acc}, the categories of the input images are limited to the target vocabulary ($(x,y)\in \mathcal{D}^{(T)}$), but CLIP is asked to distinguish all categories from a larger vocabulary $\mathcal{V}^{(T)} \cup \mathcal{V}^{(NT)}$. In other words, compared to traditional closed-set classification, CLIP is expected to reject all the negative categories from $\mathcal{V}^{(NT)}$. 
The model is required to distinguish visual concepts stably and robustly, rather than making wrong predictions in the presence of other distractors. 
Based on Eq.\eqref{eq:CLIP-condition-acc}, we define the stability of CLIP in the open task as:



\begin{figure*}[tbp]
  \centering
  \subfloat[CIFAR10]{
    \label{sfig:adv-cifar10}
    \includegraphics[width=.48\textwidth]{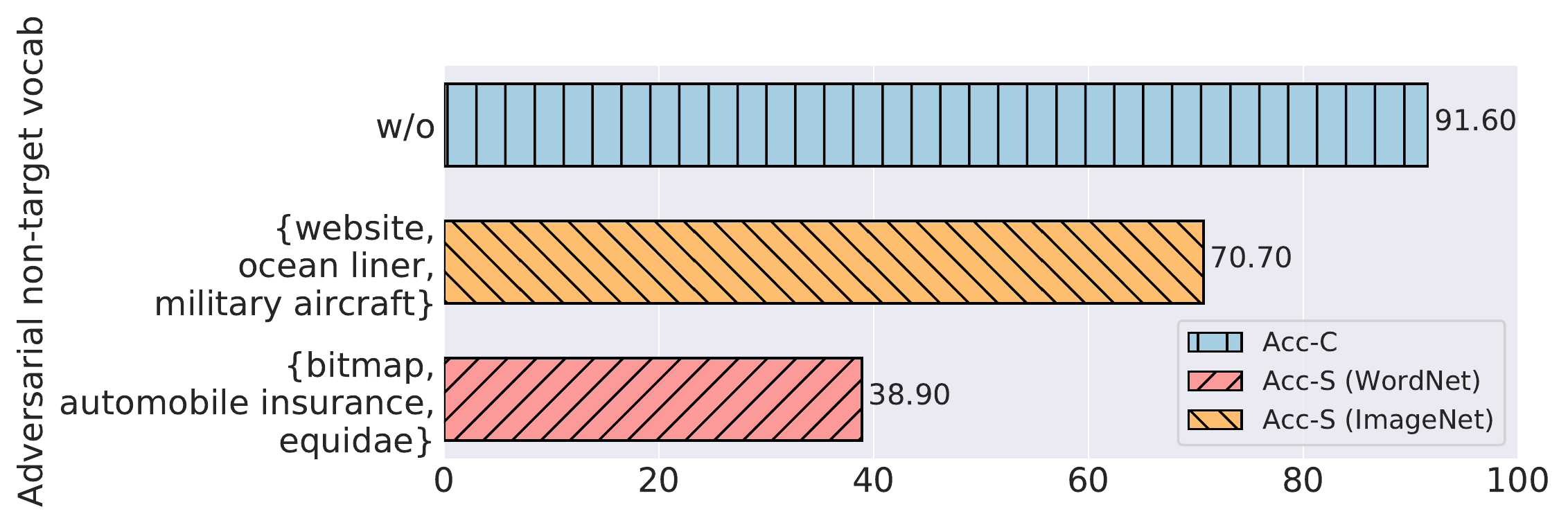}} 
  \subfloat[CIFAR100]{
    \label{sfig:adv-cifar100}
    \includegraphics[width=.45\textwidth]{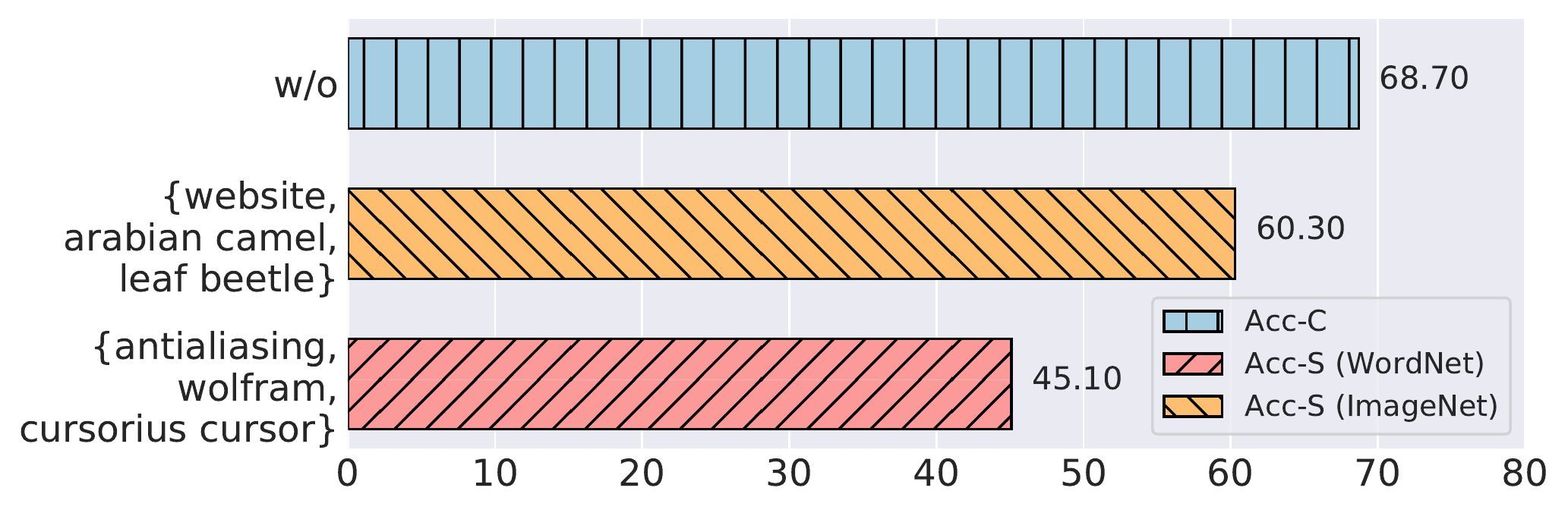}}
  \caption{Adversarial non-target vocabulary for CIFAR datasets. Adding $3$ adversarial non-target classes leads to severe performance (Acc-S) deterioration, revealing the vulnerability of CLIP when faced with malicious vocabulary.}
  \label{fig:adv}
\end{figure*}

\begin{definition}[Stability]
\label{definition:stability}
Given a target vocabulary $\mathcal{V}^{(T)}$ and $M$ non-target vocabularies $\{ \mathcal{V}^{(NT)}_1, \cdots, \mathcal{V}^{(NT)}_{M} \}$, we denote $\mathcal{S}_{M}$ as their full permutation, and $\mathcal{V}_{s_i}^{(NT)}$ as the $i^{(th)}$ vocabulary in a permutation $s \in \mathcal{S}_{M}$.
We design the \textbf{local stability} to measure the averaged classification accuracy of CLIP on the given target vocabulary when non-target vocabularies are extended incrementally:
\begin{equation}
\small
\begin{split}
& \text{Acc-}\tilde{\text{S}} \left(\mathcal{V}^{(T)}, \mathcal{V}^{(NT)} \right) = \\
& \eop_{s\in \mathcal{S}_{M}}
\frac{1}{M}\sum_{i=1}^{M} \text{Acc}\left( \mathcal{V}^{(T)} \middle| \mathcal{V}^{(T)} \cup \left( \mathcal{V}^{(NT)}_{s_1} \cup \cdots \cup \mathcal{V}^{(NT)}_{s_i} \right)\right).
\label{eq:acc-s}
\end{split}
\end{equation}
As Eq.\eqref{eq:acc-s} only reflects the local stability with respect to a single target vocabulary, we further design the \textbf{general stability} as an average of local stability over a set of target vocabularies to reduce the bias from data distribution and vocabulary sampling. 
Specifically, given $N$ vocabularies $\{\mathcal{V}_{1}, \cdots, \mathcal{V}_{N}\}$, we regard each vocabulary $\mathcal{V}_{i}$ as the target vocabulary $\mathcal{V}^{(T)}$ and the rest $\mathcal{V}_{\neq i}$ as the non-target vocabularies $\mathcal{V}^{(NT)}$, and then formulate the general stability as:
\begin{equation}
\small 
\text{Acc-S} = \frac{1}{N} \sum_{i=1}^{N} \text{Acc-}\tilde{\text{S}} \left(\mathcal{V}_{i}, \mathcal{V}_{\neq i}\right).\label{eq:general-acc-s}
\end{equation}
\end{definition}
\paragraph{Experimental settings and results}
The models and datasets adopted for evaluation are consistent with that in \textsection~\ref{subsec:extensibility}.
For the calculation of stability, take CIFAR100 with $N=20$ vocabularies as an example, we treat each vocabulary as the target vocabulary and the rest are treated as the non-target vocabularies. 
To calculate the expectation in Eq.\eqref{eq:acc-s}, we sample $100$ permutations for $M=19$ non-target vocabularies and report the averaged scores.

Table~\ref{tb:stability-extensibility} demonstrates the stability of CLIP-like models. 
On CIFAR100, the Acc-S of CLIP (RN101) decreased by $13.4\%$. 
Figure~\ref{sfig:acc-s-insects} shows Acc-S on CIFAR100 during non-target vocabulary expansion. Given a closed $\mathcal{V}^{(T)}=$ \emph{Insects}, CLIP (ViT-B/32) achieves an accuracy of $81.2\%$. However, when the remaining $19$ non-target vocabularies are incorporated, the accuracy sharply drops to $57.0\%$. The decrease of Acc-S brought by the introduction of each non-target vocabulary indicates that more images from \emph{Insects} are incorrectly classified into the new vocabulary. 
Figure~\ref{sfig:acc-o-acc-s} demonstrates the difference between Acc-C and Acc-S for each target vocabulary. When $\mathcal{V}^{(T)}=$ \emph{Medium-sized Mammals}, CLIP is most easily interfered with by the non-target vocabularies, with a $21.08\%$ performance drop. It suggests that \textbf{the unstable predictions lead to the poor extensibility of CLIP when new categories are introduced}.  
Besides, we notice that CLIP performs stably on groups like \emph{Flowers}, where its Acc-S only declines by $0.53\%$ compared to Acc-C. 
The different behaviors of different groups indicates that \textbf{the stability is also influenced by the inherent property of the image categories and naming variation}~\citep{Silberer2020ObjectNI, Takmaz2022LessDY}. 

\subsubsection{Adversarial non-target vocabulary}
\label{sec:adversarial-vocab}
In order to explore the lower bound of the stability of CLIP, we define the \textit{adversarial non-target vocabulary} $\mathcal{V}^{(ANT)}$ as the non-target vocabulary that reduces Acc-S the most:
\begin{equation}
\small 
    \mathcal{V}^{(ANT)} = \min_{\mathcal{V}^{(NT)}} \text{Acc}\left( \mathcal{V}^{(T)} \middle| \mathcal{V}^{(T)} \cup \mathcal{V}^{(NT)}  \right).
\end{equation}
To build $\mathcal{V}^{(ANT)}$, we refer to the method of adversarial examples generation~\cite{Ren2019GeneratingNL} to traverse the words in a large vocabulary, e.g., the vocabulary of nouns in  WordNet~\cite{Fellbaum2000WordNetA}, which are regarded as non-target classes in order to calculate Acc-S, and then take the most confusing words to form the adversarial non-target vocabulary. 

We constrain the size of $\mathcal{V}^{(ANT)}$ to $3$.
Results in Figure~\ref{fig:adv} illustrate the performance with nouns in WordNet and class names in ImageNet as the candidate vocabulary, respectively. First, we observe a clear performance degradation on both datasets under adversarial attack, e.g., adding \emph{bitmap}, \emph{automobile insurance} and \emph{equidae} leads to an absolute $52.7$\% accuracy drop on CIFAR10. 
Besides, we find that the selected adversarial words are much less concrete than common visual concepts like \emph{Flower}, indicating the potential reason behind is the poor semantic modeling of CLIP on those objects with higher abstraction levels.
This investigation reveals that \textbf{CLIP is vulnerable when facing malicious non-target vocabulary}, and we hope future work may pay more attention to the robustness of CLIP under open recognition tasks.

\begin{figure*}[htb]
\begin{minipage}{0.65\textwidth}
  \centering
  \includegraphics[width=\textwidth]{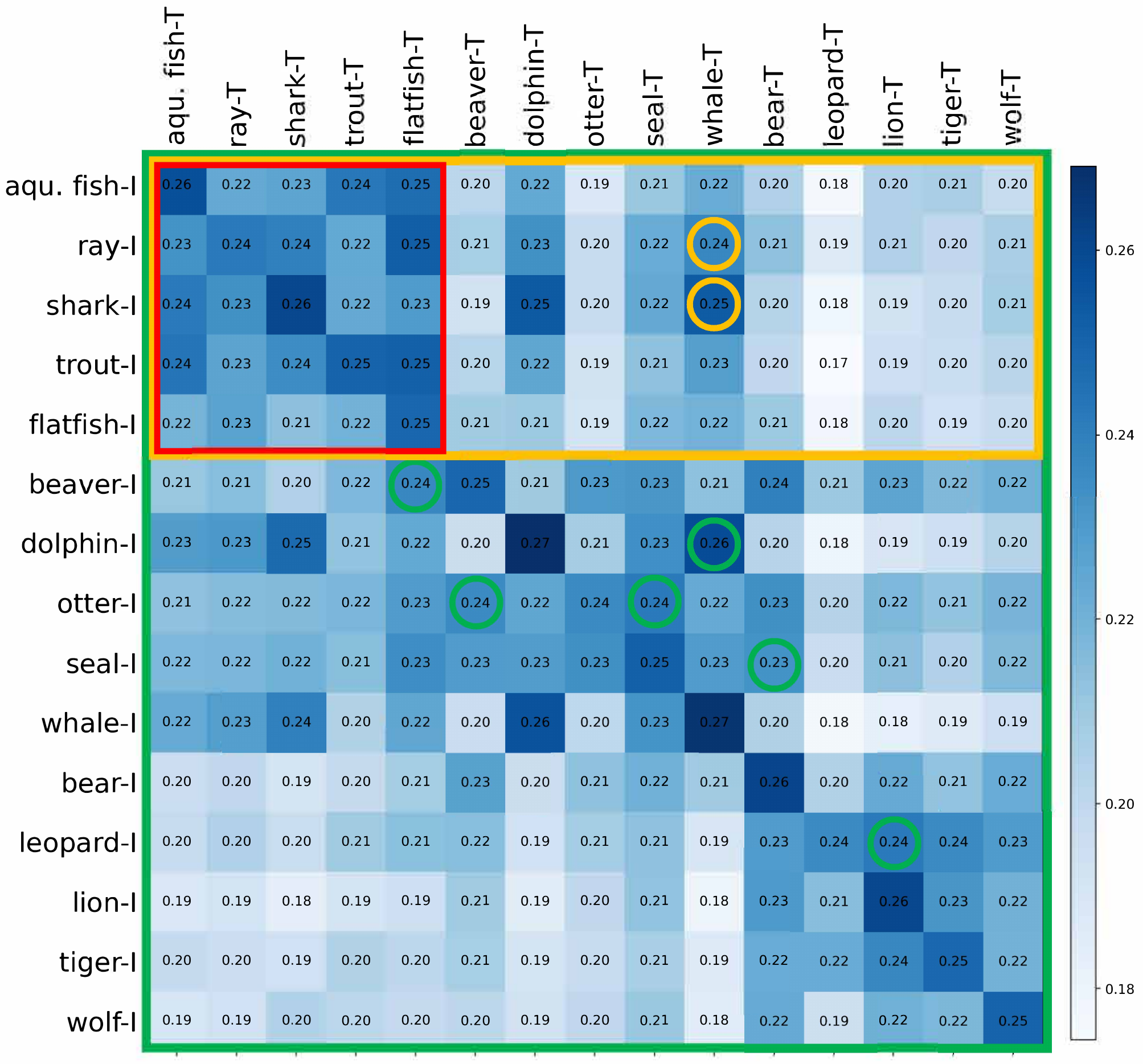}
  \caption{Cosine similarity between image~(-I) and text~(-T) features of CLIP on CIFAR100. Each value in the matrix are averaged over $100$ samples. The expansions from the red box to the green box (diagonal) and the yellow box (horizontal) refer to the calculation of extensibility and stability, respectively. The circle represents that more than $15$ wrong predictions have arisen after adding this class.}
  \label{fig:cos-sim-15-cifar100}
\end{minipage}\hfill
\begin{minipage}{0.33\textwidth}
  \centering
  \includegraphics[width=0.85\textwidth]{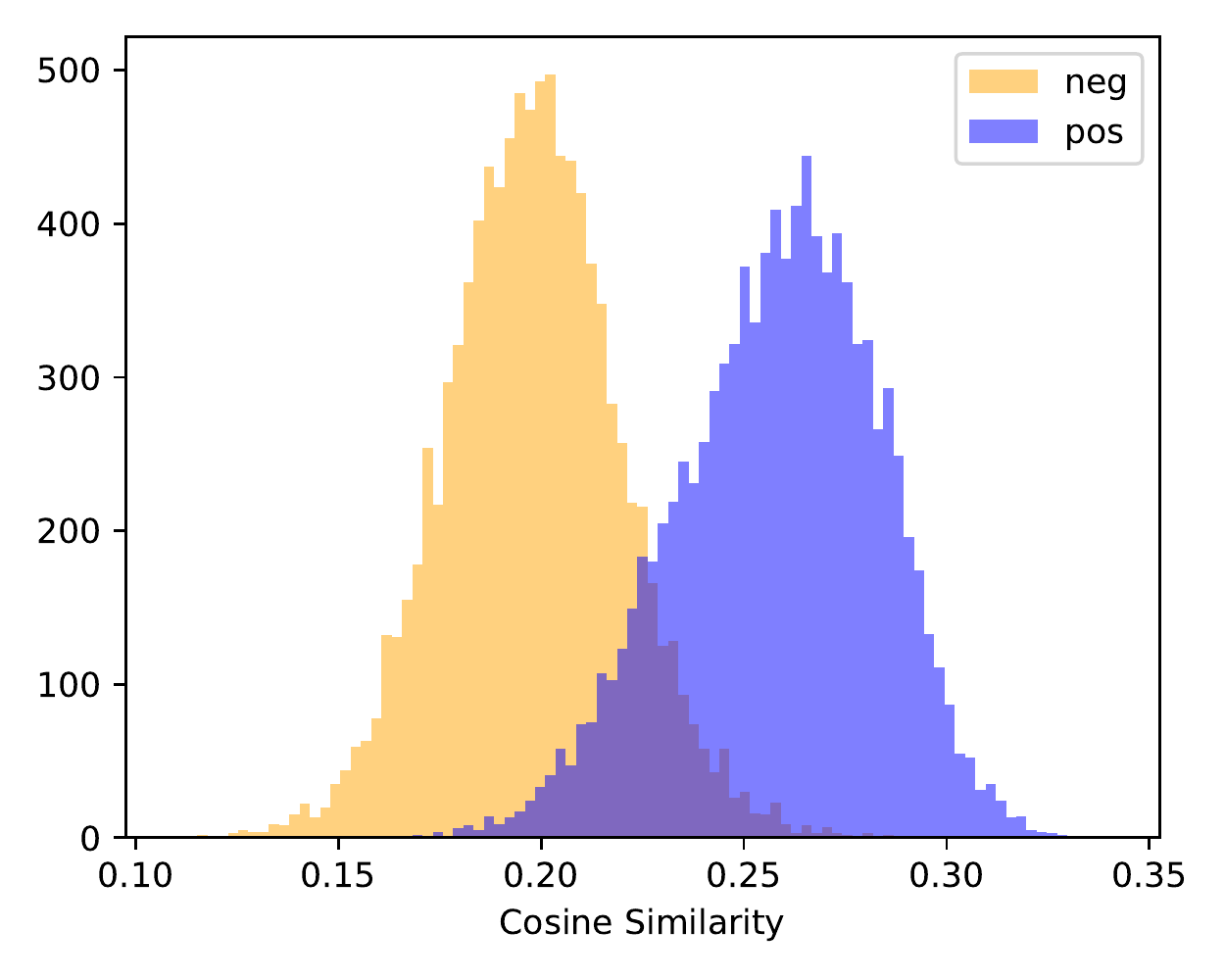}
  \caption{Cosine similarity histogram of positive (pos) and negative (neg) image-text pairs with large overlap.}
  \label{fig:cos-sim-hist-cifar100}
  \includegraphics[width=0.85\textwidth]{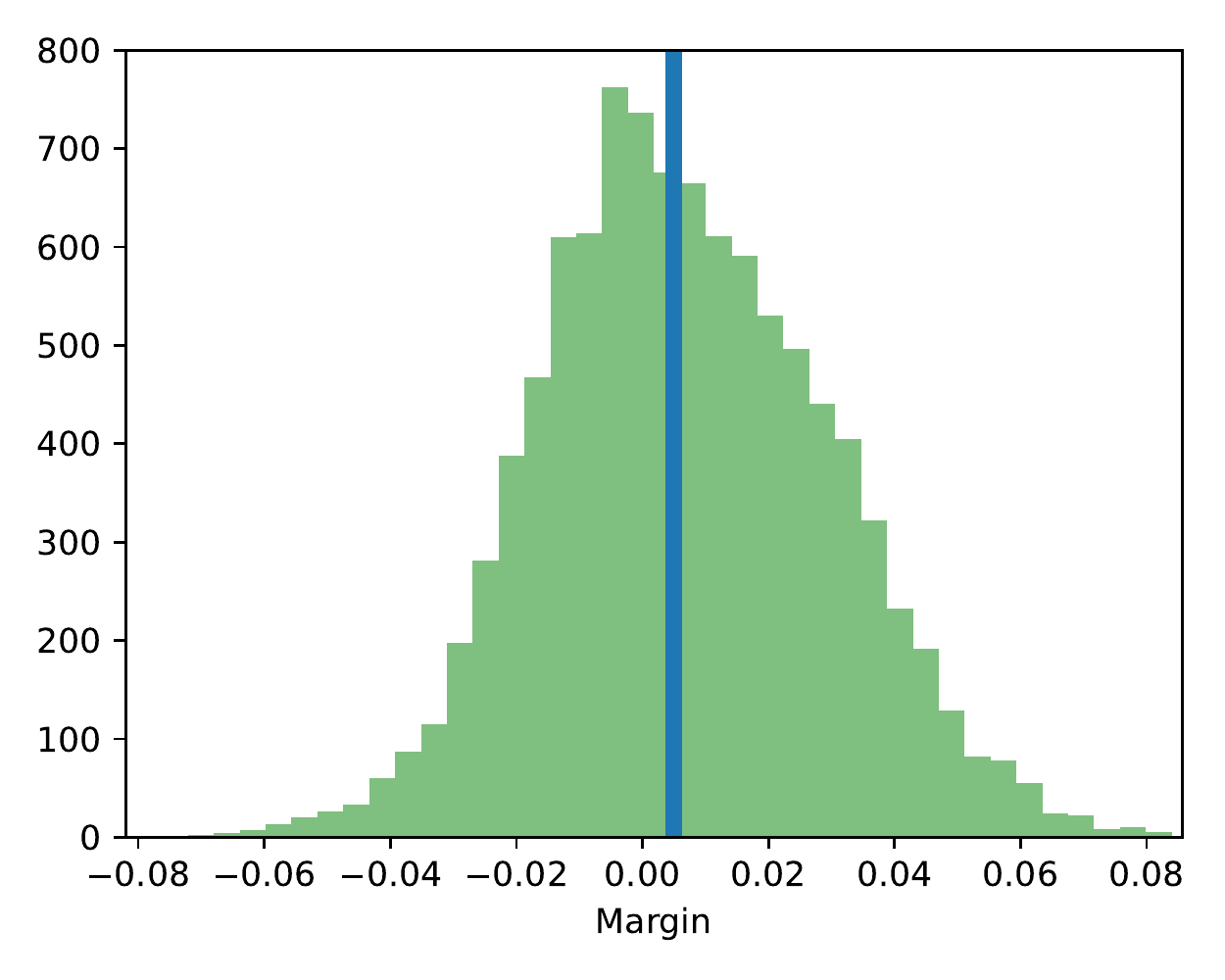}
  \caption{Margin distribution of similarity scores, which are centered around zero with a median value of $.005$~(the blue vertical line). It indicates that the predictions can be easily inverted with competing classes appearing.}
  \label{fig:cos-sim-hist-margin-cifar100}
\end{minipage}
\end{figure*}

\section{Dissecting the extensibility of CLIP}
Our experimental results in \textsection~\ref{sec:openness-extensibility-stability} reveal the poor performance of CLIP on open tasks. 
In this section, we delve into the representation space of CLIP to understand its extensibility. We first point out that the small margin between positive and negative class descriptions leads to the prediction shifting when competing class features appear, which thus limits the stability of CLIP (\textsection~\ref{subsec:small_margin}). 
Further, we investigate the representation space of CLIP-like models via two metrics: inter-modal alignment and intra-modal uniformity. The results show that enforcing the distinguishability of class features increases the margin and makes the models scale more stably (\textsection~\ref{subsec:alignment-uniformity}). 


\subsection{Small margin limits the stability of CLIP}
\label{subsec:small_margin}
Since CLIP formalizes the visual recognition as an image-to-text matching task, each text feature of the class description corresponds to the class vector in traditional classifiers, and the image-text similarity scores are analogous to the logits in classification.
Ideally, regardless of vocabulary expansion, for an image, the similarity of the positive pair (the image with the text specifying the ground-truth class) should be higher than that of the negative pairs (the image with the texts specifying other classes) to ensure the correct prediction on open tasks. In other words, the \emph{margin}~\cite{Jiang2019PredictingTG} between positive and the largest negative similarity is a direct contributor to stability.

Unfortunately, the similarity and margin distribution of CLIP do not meet our expectations. 
Figure~\ref{fig:cos-sim-15-cifar100} illustrates the averaged cosine similarity of CLIP~(ViT-B/32) on $15$ classes of CIFAR100. 
The diagonal elements represent the similarity of the positive image-text pairs, while the others represent that of the negative ones. 
In general, the cosine similarity of image-text pairs is very low, with an average of $0.20$. This number is only $0.26$ even for the positive pairs. 
Besides, the similarities of positive and negative pairs are very close, indicating the low distinguishability between different classes. 
As shown in Figure~\ref{fig:cos-sim-hist-cifar100} and Figure~\ref{fig:cos-sim-hist-margin-cifar100}, the similarity histogram of positive and negative pairs has a large overlap, and the margin is clustered around zero, leaving the predictions of models at risk of being reversed to new non-target classes. 
For example, as the vocabulary extends from the red box to the green box~(diagonal) or the yellow box~(horizontal) in Figure~\ref{fig:cos-sim-15-cifar100}, more deceptive classes (circles) with negative margins are added, leading to prediction shift. 
Particularly, the classes belonging to the same vocabulary\footnote{Every $5$ adjacent classes in Figure~\ref{fig:cos-sim-15-cifar100} constitute a vocabulary (superclass), see Table~\ref{tb:cifar100-hierarchy} in Appendix~\ref{sec:superclass-class-hierarchy}} have higher similarity and smaller margin, making them more likely to be confused with each other.


\begin{figure}[tbp]
\centering
\includegraphics[width=.8\columnwidth]{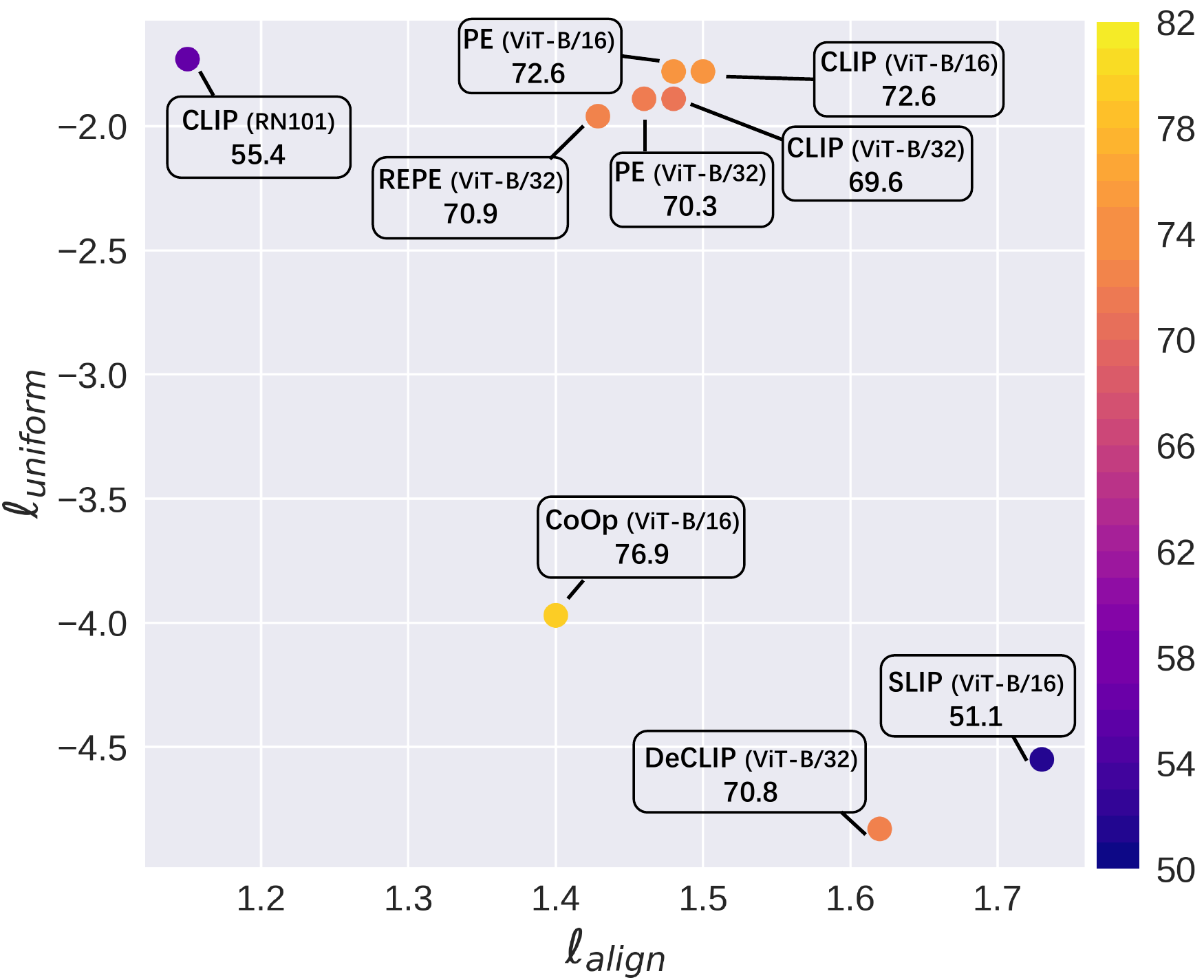}
\caption{$\ell_{\text {align}}$ and $\ell_{\text {uniform}}$ of CLIP-like models. For both two metrics, lower numbers are better. The color of points and numbers denote the extensibility performance (Acc-E) on CIFAR100 (higher is better).}
\label{fig:alignment_uniformity}
\end{figure}

\subsection{Inter-modal alignment and intra-modal uniformity ground the margin}
\label{subsec:alignment-uniformity}
\begin{figure}[t]
\centering
\includegraphics[width=.8\columnwidth]{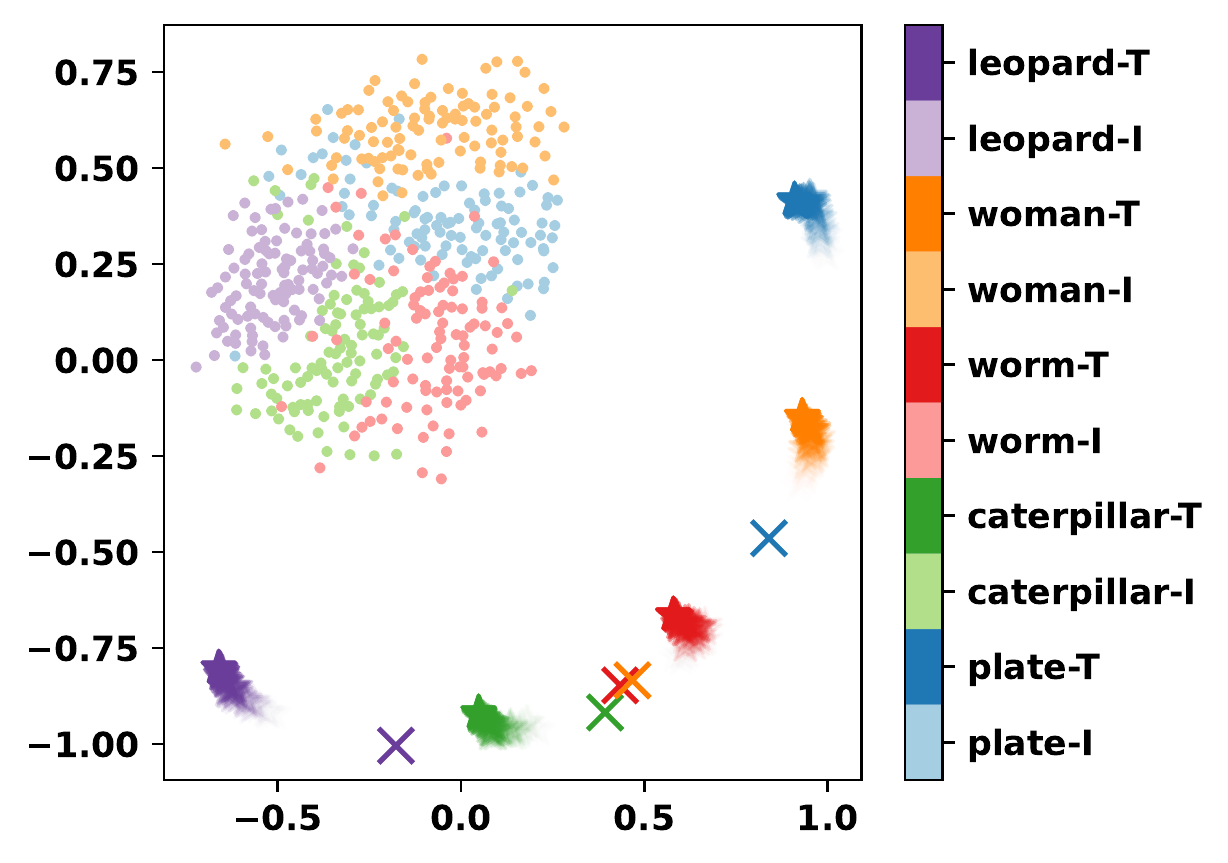}
\caption{Representation visualization of CLIP and CoOp (ViT-B/16). The five classes with different colors are from CIFAR100. $\bullet$ refers to image features (-I), while $\times$ and $\star$ refers to text features (-T) of CLIP and CoOp, respectively. The color of $\star$ from transparent to opaque indicates the optimization trajectory during the CoOp prompt-tuning process.}
\label{fig:image-text-features}
\end{figure}

According to the results in \textsection~\ref{subsec:small_margin}, the ideal feature space for CLIP-like models should have a large margin between different classes to ensure stability in open-vocabulary recognition tasks. To achieve this, the text feature of a class name should be close to the features of the images it describes~\cite{Ren2021LearningRA}, and the intra-modal features, especially textual features, should be uniformly distributed to make the descriptions of competing categories more distinguishable~\cite{Wang2020UnderstandingCR}. In order to measure the quality of representations in the vision-and-language domain, we propose two metrics, \textbf{inter-modal alignment} and \textbf{intra-modal uniformity}.
Inter-modal alignment calculates the expected distance between features of positive image-text pairs $p_{\text {pos }}$:
\begin{equation}
\small
\ell_{\text {align}} \triangleq \underset{\left(x, t\right) \sim p_{\text {pos }}}{\mathbb{E}}\left\|f_I(x)-f_T\left( t \right)\right\|^{2},
\label{eq:alignment}
\end{equation}
while intra-modal uniformity measures how well the image or text features are uniformly distributed:
\begin{equation}
\small
\begin{split}
\ell_{\text {uniform}} \triangleq & \ell_{\text {uniform-I}} + \ell_{\text {uniform-T}} \\ 
\triangleq & \log \underset{x_i, x_j \sim p_{\text {data-I}}}{\mathbb{E}} e^{-2\|f_I(x_i)-f_I(x_j)\|^{2}} + \\
& \log \underset{t_i, t_j \sim p_{\text {data-T}}}{\mathbb{E}} e^{-2\|f_T(t_i)-f_T(t_j)\|^{2}},
\label{eq:uniformity}
\end{split} 
\end{equation}
where $p_{\text {data-I}}$ and $p_{\text {data-T}}$ denotes the image and text data distribution, respectively. 
Figure~\ref{fig:alignment_uniformity} and Table~\ref{tb:alignment-uniformity} provide quantified loss of alignment and uniformity. CLIP with only cross-modal contrastive learning results in poor intra-modal uniformity ($\ell_{\text {uniform}}>-2.0$), especially on the text side. 
However, models like SLIP and DeCLIP that incorporate intra-modal contrastive learning in pre-training can better separate image and text features by classes, resulting in a much lower intra-modal uniformity loss ($\ell_{\text {uniform}}<-4.5$).
Additionally, the prompt tuning method (CoOp~\cite{Zhou2021LearningTP}) achieves better inter-modal alignment and the lowest intra-modal uniformity loss on the text side. According to the visualization via Multidimensional Scaling (MDS)~\cite{Borg1997ModernMS} in Figure~\ref{fig:image-text-features}, the optimization trajectory of prompts in CoOp leads to the cluster center of corresponding image features while also dispersing the position of prompt features, thereby improving both text uniformity and inter-modal alignment and achieving the best extensibility. 

\begin{table*}
\centering
\small 
\setlength{\tabcolsep}{3pt}
\begin{tabular}{@{}lcccccc@{}}
\toprule
\multirow{2}{*}{Model} & \multicolumn{4}{c}{Alignment \& Uniformity}                                & \multicolumn{2}{c}{Accuracy}                                    \\ \cmidrule(lr){2-5} \cmidrule(l){6-7} 
                       & $\ell_{\text {align}}$ ($\downarrow$)         & $\ell_{\text {uniform-T}}$ ($\downarrow$)      & $\ell_{\text {uniform-I}}$ ($\downarrow$)   & $\ell_{\text {uniform}}$ ($\downarrow$)   & Acc-C  ($\uparrow$)                         & Acc-E  ($\uparrow$)                          \\ \midrule
CLIP (RN101)      & \textbf{1.15} & \textbf{-1.16}             & -0.57    & -1.73                  & 68.3          & 55.4          \\
CLIP (ViT-B/32)   & 1.48                           & -0.96                      & \textbf{-0.93}      & \textbf{-1.89}       & 78.0          & 69.6          \\
CLIP (ViT-B/16)   & 1.50                           & -0.97                      & -0.81     &      -1.78           & \textbf{79.7} & \textbf{72.6} \\ \midrule
SLIP (ViT-B/16)   & 1.73                           & -2.86                      & -1.69        &      -4.55        & 63.9          & 51.1          \\
DeCLIP (ViT-B/32) & \textbf{1.62}                  & \textbf{-2.96}             & \textbf{-1.87}      &    \textbf{-4.83}       & \textbf{78.7} & \textbf{70.8} \\ \midrule
PE (ViT-B/32)     & \textbf{1.46}                  & -0.96                      & \textbf{-0.93}     &     \textbf{-1.89}     & 78.3          & 70.3          \\
PE (ViT-B/16)     & 1.48                           & \textbf{-0.97}             & -0.81      &        -1.78        & \textbf{79.6} & \textbf{72.6} \\ \midrule
CoOp (ViT-B/16)   & 1.40                           & -3.16                      & -0.81    &           -3.97       & 83.6          & 76.9          \\ \bottomrule
\end{tabular}
\caption{Inter-modal alignment ($\ell_{\text {align}}$), text uniformity ($\ell_{\text {uniform-T}}$), image uniformity ($\ell_{\text {uniform-I}}$), intra-modal uniformity ($\ell_{\text {uniform}}$), Acc-C (Eq.~\eqref{eq:acc-c}), and Acc-E (Eq.~\eqref{eq:acc-e}) of CLIP-like models on CIFAR100. For the first four metrics, lower numbers are better. For the last two metrics, higher numbers are better.}
\label{tb:alignment-uniformity}
\end{table*}

\subsection{Discussions}
After the preliminary explorations on openness of CLIP-like models, we present  potential ways to enhance the models's extensibility and stability. 

(1) For pre-training: In order to improve the quality of CLIP's feature space and enhance alignment and uniformity, more high-quality pre-training data and effective supervision signals such as $\ell_{\text {align}}$ and $\ell_{\text {uniform}}$ can be introduced during pre-training. 

(2) For zero-shot inference: Recall that in vanilla CLIP-like models, the context (hard prompt) for each class name is the same during inference, making it difficult to discriminate between distinct visual categories because the semantics of each cannot be holistically represented. To remedy this, we suggest customizing class descriptions with diverse captions retrieved from the pre-training corpus as a prompt ensemble. The effectiveness of this idea is verified through experiments, details can be found in Appendix~\ref{subsec:methodology}.

\section{Conclusion}
In this paper, we evaluate the extensibility of CLIP-like models for open-vocabulary visual recognition. Our comprehensive study reveals that as the vocabulary expands, the performance of these models deteriorates significantly due to indistinguishable text features among competing classes. We hope that our investigation and analysis will facilitate future research on the CLIP openness issue.

\section*{Limitations}
\label{sec:limitations}
To facilitate future research, we analyze the difficulties and
possible solutions in this new area. \textbf{(1)} As we present extensive empirical results and address the weakness of CLIP on vocabulary expansion, its theoretical risk on open tasks is urged to be investigated. 
\textbf{(2)} The current evaluation protocol is an approximation of the real open world. An evolving benchmark could facilitate future research. 
\textbf{(3)} For various visual categories, their degree of abstraction, the ease of describing them in natural language, and their density in the data distribution can also influence the extensibility and stability of models, which are worth studying. 

\section*{Acknowledgement}
The authors would like to thank the reviewers for their helpful comments. This work is supported by Natural Science Foundation of China (NSFC) No. 62176002. Xu Sun is the corresponding author.

\bibliography{anthology,custom}
\bibliographystyle{acl_natbib}

\appendix

\section{Appendix}
\label{sec:appendix}

\subsection{Comparison of related work}
\label{sec:related-work-comparison}
\begin{table*}[tbp]
\centering
\setlength{\tabcolsep}{3pt}
\begin{adjustbox}{max width=\textwidth}
\begin{tabular}{@{}lccccc@{}}
\toprule
Task                        & Paradigm       & Goal                                                                                                                                                                       & Signal      & Training                                                             & Testing                                                                    \\ \midrule
Closed Set Recognition      & Classification & Identifying known classes                                                                                                                                                  & Supervised   & Known classes                                                        & Known classes                                                              \\ \midrule
Open Set Recognition        & Classification & \begin{tabular}[c]{@{}c@{}}Identifying known classes \&\\ rejecting unknown classes\end{tabular}                                                                           & Supervised   & Known classes                                                        & \begin{tabular}[c]{@{}c@{}}Known classes \&\\ unknown classes\end{tabular} \\ \midrule
Open World Recognition      & Classification & \begin{tabular}[c]{@{}c@{}}Identifying known classes \&\\ detecting unknown classes \&\\ labeling unknown data \&\\ incrementally learn and extend classifier\end{tabular} & Supervised   & \begin{tabular}[c]{@{}c@{}}Incremental \\ known classes\end{tabular} & \begin{tabular}[c]{@{}c@{}}Known classes \&\\ unknown classes\end{tabular} \\ \midrule
Open-vocabulary Recognition & Matching       & Identifying classes via natural language                                                                                                                                   & Unsupervised & -                                                                    & Classes in a vocabulary                                                    \\ \bottomrule
\end{tabular}
\end{adjustbox}
\caption{A comparison of Closed Set Recognition, Open Set Recognition (OSR), Open World Recognition, and Open-vocabulary Recognition (OVR).}
\label{tb:related-work}
\end{table*}
Table~\ref{tb:related-work} provides a more detailed comparison of Closed Set Recognition (OSR)~\cite{Scheirer2013TowardOS, Geng2021RecentAI}, Open World Recognition (OWR)~\cite{Bendale2015TowardsOW}, and Open-vocabulary Recognition (OVR)~\cite{Radford2021LearningTV} from $5$ perspectives of paradigm, goal, signal, classes type in training, and classes type in testing, respectively. Contrary to the above research, CLIP-based OVR aims to identify novel classes in a zero-shot way. Since categories of images in CLIP are represented by natural language rather than discrete label IDs, CLIP can directly synthesize textual descriptions of novel classes for matching, sparing relabeling additional training data and re-training the entire model.

\subsection{Superclass-class hierarchy for vocabulary construction}
\label{sec:superclass-class-hierarchy}
To construct the vocabularies in \textsection~\ref{sec:openness-extensibility-stability}, we leverage the underlying superclass-class hierarchical structure of CIFAR100~\cite{Krizhevsky2009LearningML} and ImageNet~\cite{Deng2009ImageNetAL}, and group the classes belonging to the same superclass into a vocabulary. 
Table~\ref{tb:cifar100-hierarchy} lists the vocabularies in CIFAR100, which are specified by \cite{Krizhevsky2009LearningML}. There are $20$ vocabularies, each with $5$ classes.
For ImageNet, we utilize two superclass-class structures, Entity13 and Living17~\cite{Santurkar2021BREEDSBF}, as shown in Table~\ref{tb:entity13-hierarchy} and Table~\ref{tb:living17-hierarchy}, respectively. Entity13 has $13$ vocabularies, each with $20$ classes, while Living17 has $17$ vocabularies, each with $4$ classes.


\subsection{Dataset-level extensibility}
\label{sec:dataset-level-expansion}
The evaluation protocol in \textsection~\ref{sec:openness-extensibility-stability} estimates the extensibility and stability within a single task dataset, where the input images and classes during the vocabulary expansion come from the same data distribution.
While the protocol is only an approximation of the real open world, current CLIP-like models have exhibited serious performance degradation.
In this section, we take a step further toward real open recognition by conducting a vocabulary expansion setting at the dataset level, where the expanded vocabularies are from different datasets.
In this way, the relationship between vocabularies is more uncertain and thus can be viewed as a rigorous stress test for the CLIP-like models.
Specifically, we group all categories in a dataset into one vocabulary. Afterward, the inputs and classes of the entire new dataset are introduced at each expansion. Classes in the new vocabulary will be removed if they already exist in the previous vocabularies. 

The experiments are conducted with datasets for generic objects, including CIFAR10~\cite{Krizhevsky2009LearningML}, CIFAR100~\cite{Krizhevsky2009LearningML}, Caltech101~\cite{FeiFei2004LearningGV}, SUN397~\cite{Xiao2010SUNDL} and ImageNet~\cite{Deng2009ImageNetAL}, and specialized datasets focusing on fine-grained categories, including Flowers102~\cite{Nilsback2008AutomatedFC}, OxfordPets~\cite{Parkhi2012CatsAD} and StanfordCars~\cite{Krause20133DOR}.
Without loss of the generality, we merge $3$ datasets and evaluate the following dataset compositions:
\begin{enumerate}[label={(\arabic*)},parsep=0pt]
    \item CIFAR100-Caltech101-SUN397
    \item CIFAR10-CIFAR100-ImageNet
    \item Flowers102-OxfordPets-StanfordCars
\end{enumerate}
Composition (1) and (2) probe the performance when all the expanded datasets are generic thus the classes in different datasets are potentially semantic-correlated, while the composition (3) targets at scenarios where the coming datasets have little correlation with previous ones.
To eliminate the effect of vocabulary expansion order, we report the average performance of all $\mathrm{A}_3^3 = 6$ possible trials for each composition. 

Table~\ref{tb:dataset-stability-extensibility} demonstrates the result of the dataset-level expansion. \textbf{First}, the performance of CLIP-like models on generic dataset expansion drops dramatically. For example, the accuracy (Acc-E) of CLIP (RN101) decreases by an averaged absolute point of $14.2$ on the  \emph{CIFAR100-Caltech101-SUN397} composition during expansion, and $14.5$ on the \emph{CIFAR10-CIFAR100-ImageNet} composition. 
Due to the existence of subclass-superclass relationship for some classes in different generic datasets, e.g., \emph{cat} in CIFAR10 and \emph{tiger cat} in ImageNet, CLIP is extremely unstable on such expansion across generic datasets. For example, the Acc-S of CLIP (RN101) on the \emph{CIFAR10-CIFAR100-ImageNet} composition is 28.2\% lower than Acc-C, indicating the models are prone to be confused about the subclass-superclass relationship. 
\textbf{Meanwhile}, the CLIP-like models exhibit much better extensibility and stability on the dataset-level expansion across specialized datasets, e.g., the \emph{Flowers102-OxfordPets-StanfordCar} composition. 
The vocabularies of this composition are intrinsically disjoint in semantics, so the model can be stably extended. 
%
\textbf{In summary}, our investigations on the dataset level expansions along with the task level in the paper show the current CLIP-like models fail to meet the expectation of conducting real open vocabulary recognition.
\begin{table*}[tbp]
\centering
\setlength{\tabcolsep}{3pt}
\begin{adjustbox}{max width=.7\textwidth}
\begin{tabular}{ll}
\toprule
Vocabulary (Superclass)                     & Classes                                               \\ \midrule
aquatic                        & mammals beaver, dolphin, otter, seal, whale           \\
fish                           & aquarium fish, flatfish, ray, shark, trout            \\
flowers                        & orchids, poppies, roses, sunflowers, tulips           \\
food                           & containers bottles, bowls, cans, cups, plates         \\
fruit and vegetables           & apples, mushrooms, oranges, pears, sweet peppers      \\
household electrical devices   & clock, computer keyboard, lamp, telephone, television \\
household                      & furniture bed, chair, couch, table, wardrobe          \\
insects                        & bee, beetle, butterfly, caterpillar, cockroach        \\
large carnivores               & bear, leopard, lion, tiger, wolf                      \\
large man-made outdoor things  & bridge, castle, house, road, skyscraper               \\
large natural outdoor scenes   & cloud, forest, mountain, plain, sea                   \\
large omnivores and herbivores & camel, cattle, chimpanzee, elephant, kangaroo         \\
medium-sized mammals           & fox, porcupine, possum, raccoon, skunk                \\
non-insect invertebrates       & crab, lobster, snail, spider, worm                    \\
people                         & baby, boy, girl, man, woman                           \\
reptiles                       & crocodile, dinosaur, lizard, snake, turtle            \\
small mammals                  & hamster, mouse, rabbit, shrew, squirrel               \\
trees                          & maple, oak, palm, pine, willow                        \\
vehicles 1                     & bicycle, bus, motorcycle, pickup truck, train         \\
vehicles 2                     & lawn-mower, rocket, streetcar, tank, tractor          \\ \bottomrule
\end{tabular}
\end{adjustbox}
\caption{Superclass-class hierarchy in CIFAR100. Each superclass corresponds to a vocabulary, and each vocabulary has $5$ classes. There are $20$ kinds of vocabulary in total, specified by \cite{Krizhevsky2009LearningML}.}
\label{tb:cifar100-hierarchy}
\end{table*}
\begin{table*}[tbp]
\centering
\setlength{\tabcolsep}{3pt}
\begin{adjustbox}{max width=.8\textwidth}
\begin{tabular}{ll}
\toprule
Vocabulary (Superclass)         & Classes                                                                                                                                                                                                                                                                                                                                 \\ \midrule
garment            & \begin{tabular}[c]{@{}l@{}}trench coat, abaya, gown, poncho, military uniform, \\ jersey, cloak, bikini, miniskirt, swimming trunks, \\ lab coat, brassiere, hoopskirt, cardigan, pajama, \\ academic gown, apron, diaper, sweatshirt, sarong\end{tabular}                                                                              \\ \midrule
bird               & \begin{tabular}[c]{@{}l@{}}African grey, bee eater, coucal, American coot, indigo bunting, \\ king penguin, spoonbill, limpkin, quail, kite, \\ prairie chicken, red-breasted merganser, albatross, water ouzel, goose, \\ oystercatcher, American egret, hen, lorikeet, ruffed grouse\end{tabular}                                     \\ \midrule
reptile            & \begin{tabular}[c]{@{}l@{}}Gila monster, agama, triceratops, African chameleon, thunder snake, \\ Indian cobra, green snake, mud turtle, water snake, loggerhead, \\ sidewinder, leatherback turtle, boa constrictor, garter snake, terrapin, \\ box turtle, ringneck snake, rock python, American chameleon, green lizard\end{tabular} \\ \midrule
arthropod          & \begin{tabular}[c]{@{}l@{}}rock crab, black and gold garden spider, tiger beetle, black widow, barn spider, \\ leafhopper, ground beetle, fiddler crab, bee, walking stick, \\ cabbage butterfly, admiral, lacewing, trilobite, sulphur butterfly, \\ cicada, garden spider, leaf beetle, long-horned beetle, fly\end{tabular}          \\ \midrule
mammal             & \begin{tabular}[c]{@{}l@{}}Siamese cat, ibex, tiger, hippopotamus, Norwegian elkhound, \\ dugong, colobus, Samoyed, Persian cat, Irish wolfhound, \\ English setter, llama, lesser panda, armadillo, indri, \\ giant schnauzer, pug, Doberman, American Staffordshire terrier, beagle\end{tabular}                                      \\ \midrule
accessory          & \begin{tabular}[c]{@{}l@{}}bib, feather boa, stole, plastic bag, bathing cap, \\ cowboy boot, necklace, crash helmet, gasmask, maillot, \\ hair slide, umbrella, pickelhaube, mit- ten, sombrero, \\ shower cap, sock, running shoe, mortarboard, handkerchief\end{tabular}                                                             \\ \midrule
craft              & \begin{tabular}[c]{@{}l@{}}catamaran, speedboat, fireboat, yawl, airliner, \\ container ship, liner, trimaran, space shuttle, aircraft carrier, \\ schooner, gondola, canoe, wreck, warplane, \\ balloon, submarine, pirate, lifeboat, airship\end{tabular}                                                                             \\ \midrule
equipment          & \begin{tabular}[c]{@{}l@{}}volleyball, notebook, basketball, hand-held computer, tripod, \\ projector, barbell, monitor, croquet ball, balance beam, \\ cassette player, snorkel, horizontal bar, soccer ball, racket, \\ baseball, joystick, microphone, tape player, reflex camera\end{tabular}                                       \\ \midrule
furniture          & \begin{tabular}[c]{@{}l@{}}wardrobe, toilet seat, file, mosquito net, four-poster, \\ bassinet, chiffonier, folding chair, fire screen, shoji, \\ studio couch, throne, crib, rocking chair, dining table, \\ park bench, chest, window screen, medicine chest, barber chair\end{tabular}                                               \\ \midrule
instrument         & \begin{tabular}[c]{@{}l@{}}upright, padlock, lighter, steel drum, parking meter, \\ cleaver, syringe, abacus, scale, corkscrew, \\ maraca, saltshaker, magnetic compass, accordion, digital clock, \\ screw, can opener, odometer, organ, screwdriver\end{tabular}                                                                      \\ \midrule
man-made structure & \begin{tabular}[c]{@{}l@{}}castle, bell cote, fountain, planetarium, traffic light, \\ breakwater, cliff dwelling, monastery, prison, water tower, \\ suspension bridge, worm fence, turnstile, tile roof, beacon, \\ street sign, maze, chain-link fence, bakery, drilling platform\end{tabular}                                       \\ \midrule
wheeled vehicle    & \begin{tabular}[c]{@{}l@{}}snowplow, trailer truck, racer, shopping cart, unicycle, \\ motor scooter, passenger car, minibus, jeep, recreational vehicle, \\ jinrikisha, golfcart, tow truck, ambulance, bullet train, \\ fire engine, horse cart, streetcar, tank, Model T\end{tabular}                                                \\ \midrule
produce            & \begin{tabular}[c]{@{}l@{}}broccoli, corn, orange, cucumber, spaghetti squash, \\ butternut squash, acorn squash, cauliflower, bell pepper, fig, \\ pomegranate, mushroom, strawberry, lemon, head cabbage, \\ Granny Smith, hip, ear, banana, artichoke\end{tabular}                                                                   \\ \bottomrule
\end{tabular}
\end{adjustbox}
\caption{Superclass-class hierarchy in ImageNet (Entity13). Each superclass corresponds to a vocabulary, and each vocabulary has $20$ classes. There are $13$ kinds of vocabulary in total, specified by BREEDS~\cite{Santurkar2021BREEDSBF}.}
\label{tb:entity13-hierarchy}
\end{table*}
\begin{table*}[tbp]
\centering
\setlength{\tabcolsep}{3pt}
\begin{adjustbox}{max width=.8\textwidth}
\begin{tabular}{ll}
\toprule
Vocabulary (Superclass) & Classes                                                             \\ \midrule
salamander              & eft, axolotl, common newt, spotted salamander                       \\
turtle                  & box turtle, leatherback turtle, loggerhead, mud turtle              \\
lizard                  & whiptail, alligator lizard, African chameleon, banded gecko         \\
snake                   & night snake, garter snake, sea snake, boa constrictor               \\
spider                  & tarantula, black and gold garden spider, garden spider, wolf spider \\
grouse                  & ptarmigan, prairie chicken, ruffed grouse, black grouse             \\
parrot                  & macaw, lorikeet, African grey, sulphur-crested cockatoo             \\
crab                    & Dungeness crab, fiddler crab, rock crab, king crab                  \\
dog                     & bloodhound, Pekinese, Great Pyrenees, papillon                      \\
wolf                    & coyote, red wolf, white wolf, timber wolf                           \\
fox                     & grey fox, Arctic fox, red fox, kit fox                              \\
domestic cat            & tiger cat, Egyptian cat, Persian cat, Siamese cat                   \\
bear                    & sloth bear, American black bear, ice bear, brown bear               \\
beetle                  & dung beetle, rhinoceros beetle, ground beetle, long-horned beetle   \\
butterfly               & sulphur butterfly, admiral, cabbage butterfly, ringlet              \\
ape                     & gibbon, orangutan, gorilla, chimpanzee                              \\
monkey                  & marmoset, titi, spider monkey, howler monkey                        \\ \bottomrule
\end{tabular}
\end{adjustbox}
\caption{Superclass-class hierarchy in ImageNet (Living17). Each superclass corresponds to a vocabulary, and each vocabulary has $4$ classes. There are $17$ kinds of vocabulary in total, specified by BREEDS~\cite{Santurkar2021BREEDSBF}.}
\label{tb:living17-hierarchy}
\end{table*}
\begin{table*}[tbp]
\centering
\begin{adjustbox}{max width=\textwidth}
\begin{tabular}{@{}lccccccccccccccc@{}}
\toprule
                        & \multicolumn{5}{c}{CIFAR100-Caltech101-SUN397}                                                                                                                                            & \multicolumn{5}{c}{CIFAR10-CIFAR100-ImageNet}                                                                                                                                             & \multicolumn{5}{c}{Flowers102-OxfordPets-StanfordCars}                                                                                                                               \\ \cmidrule(l){2-6} \cmidrule(l){7-11} \cmidrule(l){12-16} 
                        &                         & \multicolumn{2}{c}{\cellcolor[HTML]{E2F0D9}Extensibility}                      & \multicolumn{2}{c}{\cellcolor[HTML]{DAE3F3}Stability}                          &                         & \multicolumn{2}{c}{\cellcolor[HTML]{E2F0D9}Extensibility}                      & \multicolumn{2}{c}{\cellcolor[HTML]{DAE3F3}Stability}                          &                         & \multicolumn{2}{c}{\cellcolor[HTML]{E2F0D9}Extensibility}                     & \multicolumn{2}{c}{\cellcolor[HTML]{DAE3F3}Stability}                         \\ \cmidrule(lr){3-4} \cmidrule(lr){5-6} \cmidrule(lr){8-9} \cmidrule(lr){10-11} \cmidrule(lr){13-14} \cmidrule(lr){15-16} 
\multirow{-3}{*}{Model} & \multirow{-2}{*}{Acc-C} & \cellcolor[HTML]{E2F0D9}Acc-E         & \cellcolor[HTML]{E2F0D9}$\Delta$       & \cellcolor[HTML]{DAE3F3}Acc-S         & \cellcolor[HTML]{DAE3F3}$\Delta$       & \multirow{-2}{*}{Acc-C} & \cellcolor[HTML]{E2F0D9}Acc-E         & \cellcolor[HTML]{E2F0D9}$\Delta$       & \cellcolor[HTML]{DAE3F3}Acc-S         & \cellcolor[HTML]{DAE3F3}$\Delta$       & \multirow{-2}{*}{Acc-C} & \cellcolor[HTML]{E2F0D9}Acc-E         & \cellcolor[HTML]{E2F0D9}$\Delta$      & \cellcolor[HTML]{DAE3F3}Acc-S         & \cellcolor[HTML]{DAE3F3}$\Delta$      \\ \midrule
CLIP (RN101)            & 65.9                    & \cellcolor[HTML]{E2F0D9}51.7          & \cellcolor[HTML]{E2F0D9}-14.2          & \cellcolor[HTML]{DAE3F3}52.7          & \cellcolor[HTML]{DAE3F3}-13.2          & 62.4                    & \cellcolor[HTML]{E2F0D9}47.9          & \cellcolor[HTML]{E2F0D9}\textbf{-14.5} & \cellcolor[HTML]{DAE3F3}34.2          & \cellcolor[HTML]{DAE3F3}\textbf{-28.2} & 65.8                    & \cellcolor[HTML]{E2F0D9}63.1          & \cellcolor[HTML]{E2F0D9}-2.7          & \cellcolor[HTML]{DAE3F3}65.7          & \cellcolor[HTML]{DAE3F3}-0.1          \\
CLIP (ViT-B/32)         & 72.0                    & \cellcolor[HTML]{E2F0D9}59.4          & \cellcolor[HTML]{E2F0D9}\textbf{-12.6} & \cellcolor[HTML]{DAE3F3}61.2          & \cellcolor[HTML]{DAE3F3}\textbf{-10.8} & 70.9                    & \cellcolor[HTML]{E2F0D9}52.7          & \cellcolor[HTML]{E2F0D9}-18.2          & \cellcolor[HTML]{DAE3F3}41.3          & \cellcolor[HTML]{DAE3F3}-29.6          & 65.8                    & \cellcolor[HTML]{E2F0D9}62.0          & \cellcolor[HTML]{E2F0D9}-3.8          & \cellcolor[HTML]{DAE3F3}65.8          & \cellcolor[HTML]{DAE3F3}\textbf{-0.0} \\
CLIP (ViT-B/16)         & \textbf{74.6}           & \cellcolor[HTML]{E2F0D9}\textbf{60.6} & \cellcolor[HTML]{E2F0D9}-14.0          & \cellcolor[HTML]{DAE3F3}\textbf{61.7} & \cellcolor[HTML]{DAE3F3}-12.9          & \textbf{74.7}           & \cellcolor[HTML]{E2F0D9}\textbf{56.6} & \cellcolor[HTML]{E2F0D9}-18.0          & \cellcolor[HTML]{DAE3F3}\textbf{43.3} & \cellcolor[HTML]{DAE3F3}-31.4          & \textbf{72.3}           & \cellcolor[HTML]{E2F0D9}\textbf{69.6} & \cellcolor[HTML]{E2F0D9}\textbf{-2.7} & \cellcolor[HTML]{DAE3F3}\textbf{72.3} & \cellcolor[HTML]{DAE3F3}\textbf{-0.0} \\ \midrule
SLIP (ViT-B/16)         & 58.6                    & \cellcolor[HTML]{E2F0D9}44.4          & \cellcolor[HTML]{E2F0D9}-14.2          & \cellcolor[HTML]{DAE3F3}46.3          & \cellcolor[HTML]{DAE3F3}-12.3          & 55.6                    & \cellcolor[HTML]{E2F0D9}36.7          & \cellcolor[HTML]{E2F0D9}-18.9          & \cellcolor[HTML]{DAE3F3}30.5          & \cellcolor[HTML]{DAE3F3}\textbf{-25.1} & 35.0                    & \cellcolor[HTML]{E2F0D9}26.0          & \cellcolor[HTML]{E2F0D9}-9.0          & \cellcolor[HTML]{DAE3F3}35.0          & \cellcolor[HTML]{DAE3F3}\textbf{-0.0} \\
DeCLIP (ViT-B/32)       & \textbf{74.3}           & \cellcolor[HTML]{E2F0D9}\textbf{60.8} & \cellcolor[HTML]{E2F0D9}\textbf{-13.5} & \cellcolor[HTML]{DAE3F3}\textbf{63.3} & \cellcolor[HTML]{DAE3F3}\textbf{-11.0} & \textbf{73.0}           & \cellcolor[HTML]{E2F0D9}\textbf{55.4} & \cellcolor[HTML]{E2F0D9}\textbf{-17.6} & \cellcolor[HTML]{DAE3F3}\textbf{45.1} & \cellcolor[HTML]{DAE3F3}-27.9          & \textbf{70.2}           & \cellcolor[HTML]{E2F0D9}\textbf{63.3} & \cellcolor[HTML]{E2F0D9}\textbf{-6.9} & \cellcolor[HTML]{DAE3F3}\textbf{70.2} & \cellcolor[HTML]{DAE3F3}\textbf{-0.0} \\ \midrule
PE (ViT-B/32)           & 71.8                    & \cellcolor[HTML]{E2F0D9}59.9          & \cellcolor[HTML]{E2F0D9}\textbf{-11.9} & \cellcolor[HTML]{DAE3F3}59.6          & \cellcolor[HTML]{DAE3F3}\textbf{-12.2} & 72.2                    & \cellcolor[HTML]{E2F0D9}53.5          & \cellcolor[HTML]{E2F0D9}\textbf{-18.7} & \cellcolor[HTML]{DAE3F3}\textbf{41.6} & \cellcolor[HTML]{DAE3F3}\textbf{-30.6} & 65.7                    & \cellcolor[HTML]{E2F0D9}62.0          & \cellcolor[HTML]{E2F0D9}-3.7          & \cellcolor[HTML]{DAE3F3}65.7          & \cellcolor[HTML]{DAE3F3}\textbf{-0.0} \\
PE (ViT-B/16)           & \textbf{75.0}           & \cellcolor[HTML]{E2F0D9}\textbf{61.5} & \cellcolor[HTML]{E2F0D9}-13.5          & \cellcolor[HTML]{DAE3F3}\textbf{62.5} & \cellcolor[HTML]{DAE3F3}-12.5          & \textbf{75.4}           & \cellcolor[HTML]{E2F0D9}\textbf{56.7} & \cellcolor[HTML]{E2F0D9}\textbf{-18.7} & \cellcolor[HTML]{DAE3F3}41.3          & \cellcolor[HTML]{DAE3F3}-34.1          & \textbf{72.5}           & \cellcolor[HTML]{E2F0D9}\textbf{70.0} & \cellcolor[HTML]{E2F0D9}\textbf{-2.5} & \cellcolor[HTML]{DAE3F3}\textbf{72.5} & \cellcolor[HTML]{DAE3F3}\textbf{-0.0} \\ \bottomrule
\end{tabular}
\end{adjustbox}
\caption{Extensibility and stability of CLIP and its variants during dataset-level vocabulary expansion. 
$\Delta$ refers to the decline of Acc-E/Acc-S (\%) compared to Acc-C (\%). PE denotes Prompt Ensemble.} 
\label{tb:dataset-stability-extensibility}
\end{table*}

\subsection{Incremental Acc-E and Acc-S on CIFAR100}
\label{sec:incremental-acc}
We record the Acc-E (Eq.\eqref{eq:acc-e}) and Acc-S (Eq.\eqref{eq:acc-s}) after each vocabulary expansion on CIFAR100 to investigate the openness of CLIP-like models.

\begin{figure*}[htb]
    \centering
    \includegraphics[width=0.95\textwidth]{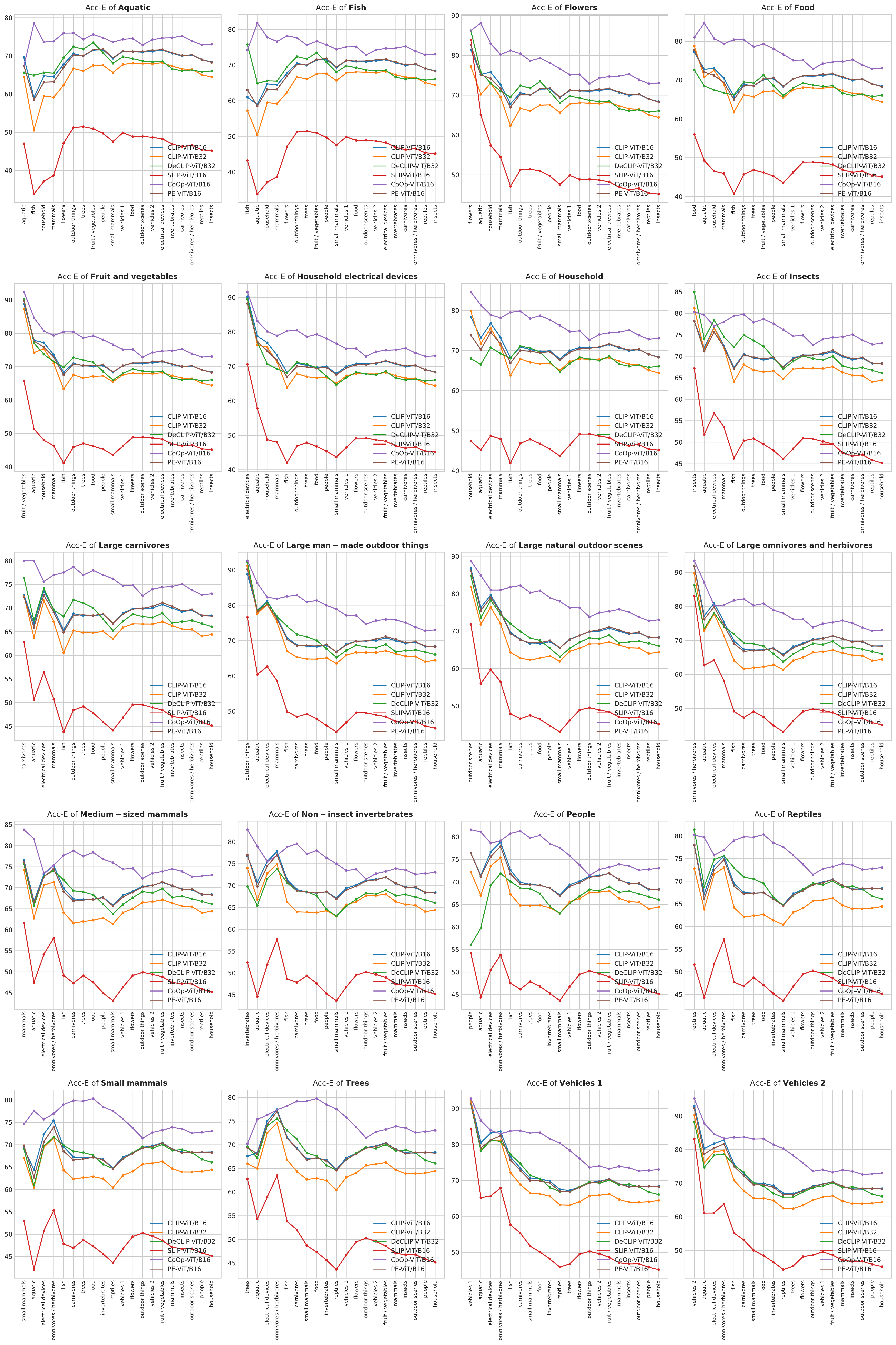}
    \caption{Incremental Acc-E of CLIP and its variants on CIFAR100.}
    \label{fig:acc-e-cifar100}
\end{figure*}

Figure~\ref{fig:acc-e-cifar100} shows the Acc-E for $20$ trials as new vocabularies are merged incrementally. The falling lines indicate that the model is either performing poorly on the new input images, or that some images that were correctly identified before are misclassified after introducing the new classes.

\begin{figure*}[htb]
    \centering
    \includegraphics[width=0.95\textwidth]{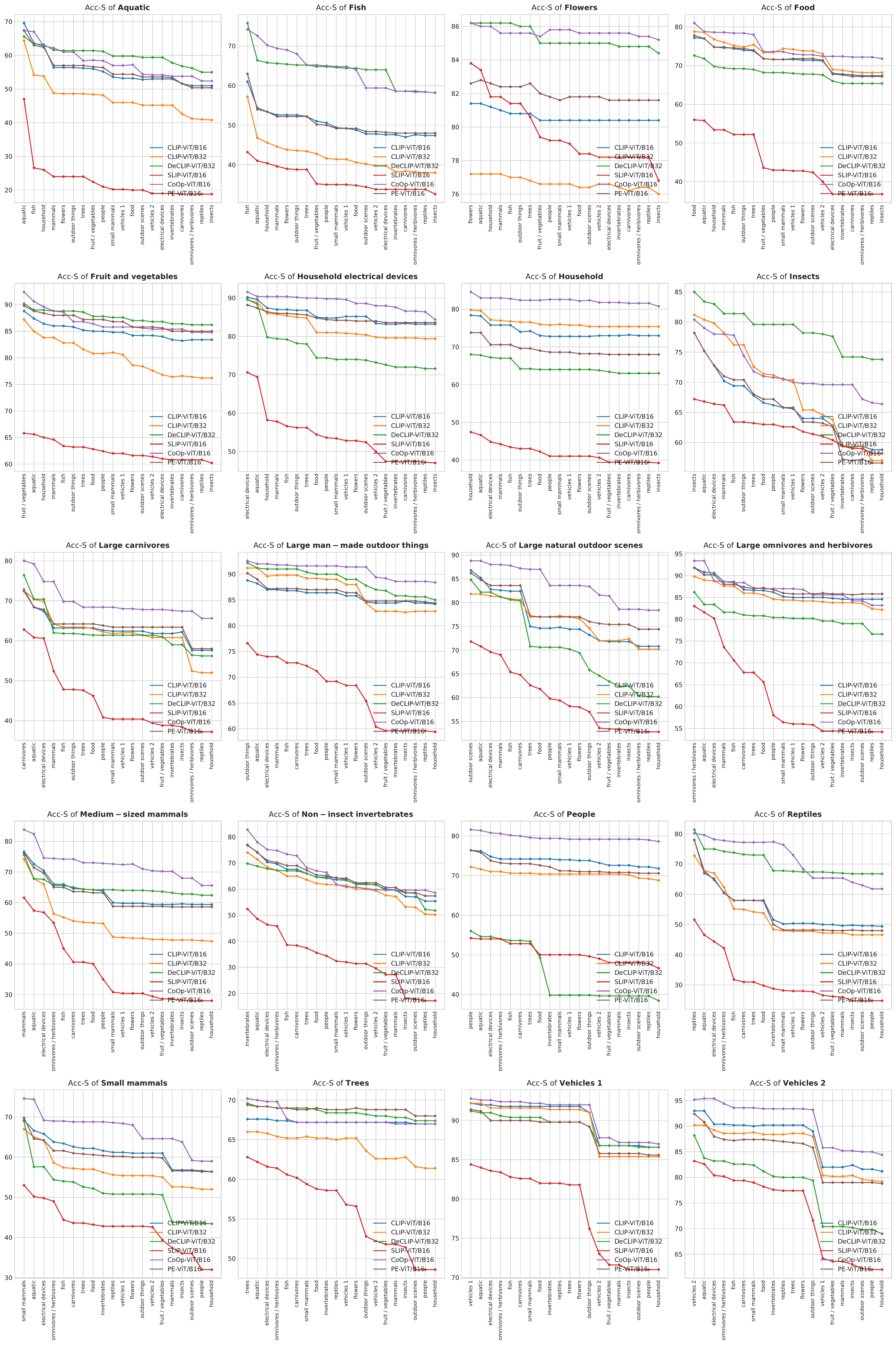}
    \caption{Incremental Acc-S of CLIP and its variants on CIFAR100.}
    \label{fig:acc-s-cifar100}
\end{figure*}

Figure~\ref{fig:acc-s-cifar100} shows Acc-S of CLIP-like models during non-target vocabulary expansion. Each sub-figure represents the situation when one vocabulary is selected as the target vocabulary. As the remaining $19$ non-target vocabularies are incorporated and the model is required to recognize the $5$ target classes from $100$ potential classes, the accuracy drops sharply. The decrease of Acc-S brought by each introduction of non-target vocabulary indicates that more images from the target vocabulary are incorrectly classified into the new non-target vocabulary by models. 

\subsection{Retrieval-enhanced prompt engineering}
\label{subsec:methodology}
In light of the previous investigations, we propose a simple yet effective method named Retrieval-enhanced Prompt Engineering (REPE) to enforce the distinguishability of class features and the image-class semantic alignment~\cite{Cao2020BehindTS, Ren2021LearningRA}. 
Recall that the context for each class name is the same in vanilla CLIP-like models (e.g., ``\texttt{a photo of a [CLASSNAME]}''), making it difficult to discriminate between distinct visual categories because the semantics of each cannot be holistically represented~\cite{Zhou2022ConditionalPL}.

To remedy this, we propose to customize each class description with diverse captions retrieved from the pre-training corpus as a prompt ensemble.
Specifically, for each class description based on the original prompt, we utilize CLIP to recall the most similar images from the pre-training dataset via image-text similarity, then obtain their corresponding captions. 
The retrieved captions with no appearance of the class name are filtered out, yielding $K$ captions. 
Such a workflow leverages both visual semantics and class names, achieving better performance. 
\begin{table*}[tbp]
\centering
\setlength{\tabcolsep}{3pt}
\begin{adjustbox}{max width=.7\textwidth}
\begin{tabular}{@{}ll@{}}
\toprule
Class                   & Retrieved captions                                               \\ \midrule
\multirow{3}{*}{apple}  & ``Apple slices stacked on top of each other''                      \\
                        & ``Apples growing on a tree''                                       \\
                        & ``Still life with apples in a basket''                             \\ \midrule
\multirow{3}{*}{woman}  & ``Portrait of a young woman''                                      \\
                        & ``Woman standing at the window''                                   \\
                        & ``Confident woman in a red dress and gold crown''                  \\ \midrule
\multirow{3}{*}{bridge} & ``The golden bridge in Bangkok''                                   \\
                        & ``Bridge on the River Kwai $\sim$Video Clip''                      \\
                        & ``Wooden bridge over a mountain river''                            \\ \midrule
\multirow{3}{*}{ray}    & ``Stingray in the Grand Cayman, Cayman Islands stock photography'' \\
                        & ``Common Stingray swimming close to the sea floor.''               \\
                        & ``Sun Rays Tours: Go Pro captured the rays under water''           \\ \bottomrule
\end{tabular}
\end{adjustbox}
\caption{Instances of the captions retrieved by our REPE on CIFAR100.}
\label{tb:retrieved-case}
\end{table*}
\begin{table*}[tbp]
\centering
\begin{adjustbox}{max width=\textwidth}
\begin{tabular}{@{}llllllllll@{}}
\toprule
                        & \multicolumn{3}{c}{CIFAR100}                                       & \multicolumn{3}{c}{ImageNet (Entity13)}                            & \multicolumn{3}{c}{ImageNet (Living17)}                            \\ \cmidrule(l){2-4} \cmidrule(l){5-7} \cmidrule(l){8-10} 
\multirow{-2}{*}{Model} & Acc-C                & Acc-E                & Acc-S                & Acc-C                & Acc-E                & Acc-S                & Acc-C                & Acc-E                & Acc-S                \\ \midrule
CLIP (RN101)            & 68.3                 & 55.4                 & 54.9                 & 80.4                 & 77.4                 & 77.3                 & 77.6                 & 74.5                 & 74.4                 \\
\rowcolor[HTML]{EFEFEF} 
REPE (RN101)                  & $\textbf{68.4}_\impro{0.1}$                        & $\textbf{55.5}_\impro{0.1}$ & $\textbf{55.2}_\impro{0.3}$ & $\textbf{81.7}_\impro{1.3}$        & $\textbf{79.2}_\impro{1.8}$ & $\textbf{79.0}_\impro{1.7}$ & $\textbf{77.8}_\impro{0.2}$ & $\textbf{75.3}_\impro{0.8}$ & $\textbf{75.2}_\impro{0.8}$ \\

CLIP (ViT-B/32)         & 78.0                 & 69.6                 & 68.9                 & 80.8                 & 78.0                 & 77.8                 & 78.0                 & 74.4                 & 75.0                 \\
\rowcolor[HTML]{EFEFEF} 
REPE (ViT-B/32)                  & $\textbf{78.5}_\impro{0.5}$ & $\textbf{70.9}_\impro{1.3}$ & $\textbf{70.6}_\impro{1.7}$ & $\textbf{82.3}_\impro{1.5}$ & $\textbf{79.8}_\impro{1.8}$ & $\textbf{79.6}_\impro{1.8}$ & $\textbf{79.0}_\impro{1.0}$ & $\textbf{76.4}_\impro{2.0}$ & $\textbf{76.2}_\impro{1.2}$ \\
CLIP (ViT-B/16)         & 79.7                 & 72.6                 & 72.0                 & 83.5                 & 81.1                 & 81.0                 & 79.5                 & 77.9                 & 77.6                 \\
\rowcolor[HTML]{EFEFEF} 
REPE (ViT-B/16)                  & $\textbf{79.8}_\impro{0.1}$ & $\textbf{72.9}_\impro{0.3}$ & $\textbf{72.6}_\impro{0.6}$ & $\textbf{85.4}_\impro{1.9}$ & $\textbf{83.3}_\impro{2.2}$ & $\textbf{83.2}_\impro{2.2}$ & $\textbf{79.9}_\impro{0.4}$ & $\textbf{78.4}_\impro{0.5}$ & $\textbf{78.2}_\impro{0.6}$ \\ \bottomrule
\end{tabular}
\end{adjustbox}
\caption{Extensibility and stability of our REPE method on CIFAR100 and ImageNet datasets.}
\label{tb:REPE}
\end{table*}
Table~\ref{tb:retrieved-case} shows some cases of the captions retrieved by our proposed REPE on CIFAR100. They share the same target of interest with the original prompt, i.e., ``\texttt{a photo of a [CLASS]}'', but provide the context in which the class name is located and thus have richer semantics. For example, given a class like \emph{bridge}, the retrieved captions describe its possible properties (e.g., ``golden'', ``wooded''), connections to other objects (e.g., ``over a mountain river''), etc., yielding more expressive and distinguishable text features of the class. 

After retrieval, we encode the retrieved captions and conduct a mean pooling operation among them. The final text representation is: 
\begin{equation*}
\small 
f_T^\text{REPE}( t_i) = (1-\lambda) f_T( t_i) + \lambda \frac{1}{K} \sum_{j}^{K} f_T( rt_{ij}),
\end{equation*}
where $rt_{ij}$ is the $j^{(th)}$ retrieved caption for class $i$ and $\lambda$ is a weighting factor.
After that, the ensemble text representation $f_T^\text{REPE}(t_i)$ is adopted as the class anchor for conducting the image classification.
With REPE, the representation of the class description shifts towards that of the representative captions in the pre-training dataset, which alleviates the semantic inconsistency between pre-training and inference. 

\begin{table}[tbp]
\centering
\setlength{\tabcolsep}{3pt}
\begin{adjustbox}{max width=\columnwidth}
\begin{tabular}{@{}lcll@{}}
\toprule
Method              & K-shot & CIFAR100      & ImageNet      \\ \midrule
CLIP-Adapter        & 4      & 66.6          & 63.0          \\
\rowcolor[HTML]{EFEFEF} 
CLIP-Adapter + REPE & 4      & $\textbf{67.5}_\impro{0.9}$ & $\textbf{63.3}_\impro{0.3}$ \\
CLIP-Adapter        & 16     & 69.0          & 64.6          \\
\rowcolor[HTML]{EFEFEF} 
CLIP-Adapter + REPE & 16     & $\textbf{69.8}_\impro{0.8}$ & $\textbf{64.9}_\impro{0.3}$ \\ \bottomrule
\end{tabular}
\end{adjustbox}
\caption{Accuracy of CLIP-Adapter and our REPE method with few-shot learning.}
\label{tb:REPE-adapter}
\end{table}

\paragraph{Experiments}
We retrieve the images and captions from CC12M~\cite{Changpinyo2021Conceptual1P}, a subset of the pre-training dataset of CLIP. 
The images and captions are pre-encoded within \textbf{an hour} using a single RTX TITAN GPU, then we build their indices for KNN search with the FAISS framework~\cite{faiss}, which also takes about \textbf{an hour}. 
Once the indices are built, we can efficiently search over the dataset according to the query image in less than \textbf{5 ms}, which is applicable for query-intensive scenarios. 

\begin{figure}[]
\centering
\includegraphics[width=.8\columnwidth]{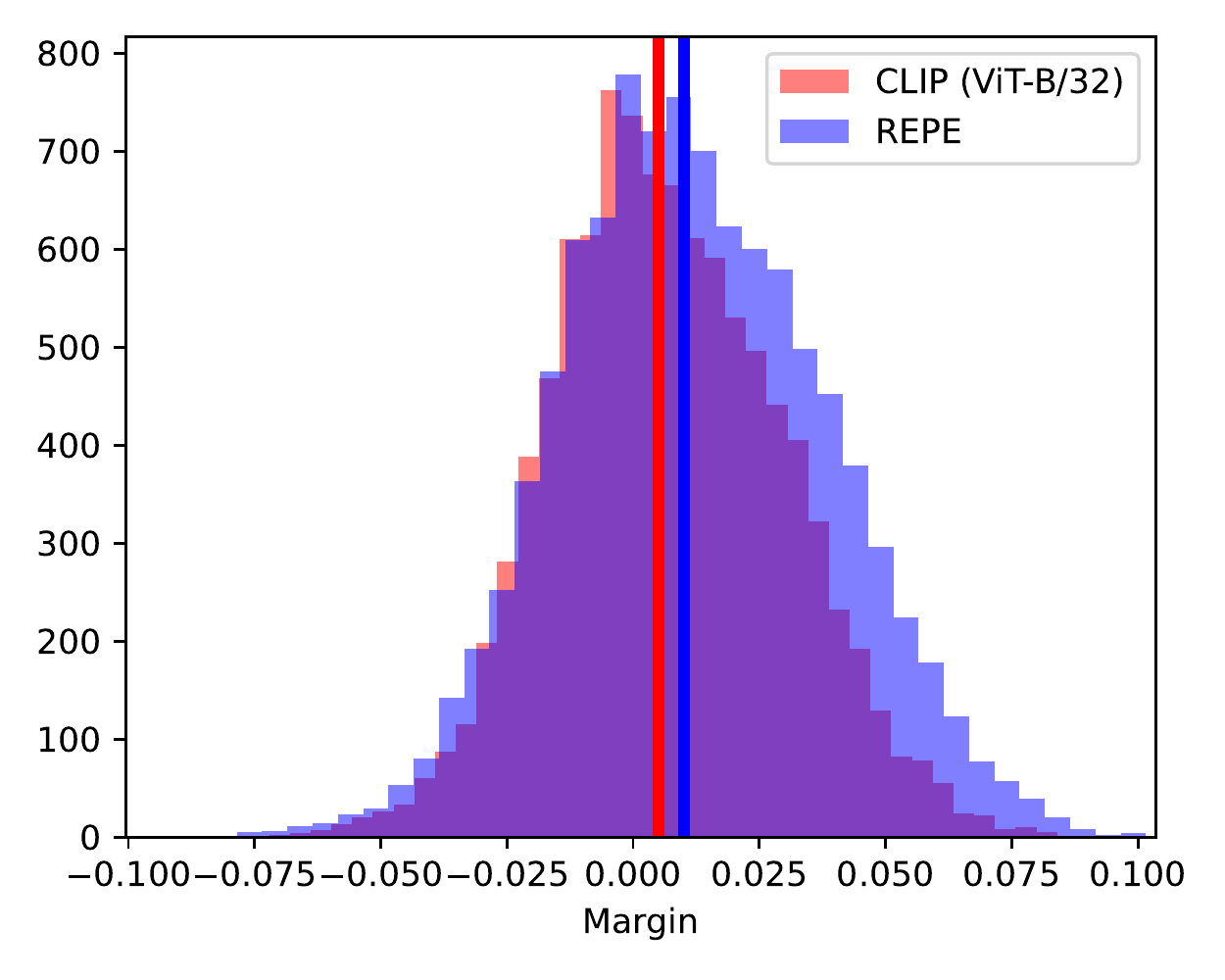}
\caption{Margin distribution of similarity scores of our REPE (blue) and CLIP (ViT-B/32) (red). The median value of REPE's distribution (the blue vertical line) is larger than that of CLIP (the red line), indicating that the predictions of REPE are harder to be inverted with competing classes than the original CLIP.}
\label{fig:REPE-margin-cifar100}
\end{figure}

Table~\ref{tb:REPE} shows the results of REPE. The hyper-parameter $K$ is $100$ and $\lambda$ is 0.25. 
REPE consistently improves the extensibility and stability of CLIP by an average of $\textbf{1.2\%}$ across all three datasets. 
We further evaluate the quality of the enhanced representations by analyzing the loss of text uniformity and inter-modal alignment. 
As shown in Figure~\ref{fig:alignment_uniformity}, our proposal effectively reduces $\ell_{\text {uniform-T}}$ from $-0.8$ to $-1.0$ and $\ell_{\text {align}}$ from $1.5$ to $1.4$, verifying its effectiveness in improving the class anchor for better extensibility and stability.
Additionally, as shown in Figure~\ref{fig:REPE-margin-cifar100}, REPE increases the median value of the margin distribution from 0.005 to 0.01 and pushes the overall distribution towards the positive side compared to vanilla CLIP. It indicates that REPE widens the gap between positive and negative class features, making it more difficult to invert predictions with competing classes. These findings support REPE's effectiveness in alleviating the openness issue. 

It is worth noting that compared to the method that requires computation-intensive pre-training procedures (DeCLIP and SLIP), and the prompt-tuning approach (CoOp) demands access to the downstream target dataset, our REPE is a lightweight framework for the zero-shot inference stage without fine-tuning. 
Besides, since REPE is model-agnostic and orthogonal to parameter-tuning methods, it can also be combined with fine-tuning methods like adapter-tuning~\cite{Gao2021CLIPAdapterBV}, to achieve a further performance boost of $0.6$ on CIFAR100 and ImageNet, which demonstrates the adaptability and superiority of our method. Please refer to Table~\ref{tb:REPE-adapter} for details. 

\end{document}